\documentclass[nocopyright,nolists]{asmejour}

\usepackage{multirow}
\usepackage{placeins}
\usepackage{float}

\newcommand{\ours}{DeepJEB++}

\newif\ifblind
\ifdefined\blindsubmission\blindtrue\else\blindfalse\fi

\JourName{Mechanical Design}

\begin{document}

\ifblind
\SetAuthorBlock{Anonymous Author(s)}{%
	Affiliation(s) omitted for double-blind review
}
\else
\SetAuthorBlock{Soyoung Yoo\textsuperscript{*}\\ Leekyo Jeong\textsuperscript{*}\\ Jinsu Ra\\ Dongeon Lee}{%
	Cho Chun Shik Graduate School of Mobility,\\
	Korea Advanced Institute of Science and Technology,\\
	Daejeon 34051, Republic of Korea
}
\SetAuthorBlock{Sunwoong Yang}{%
	Assistant Professor\\
	Department of Mechanical Engineering,\\
	Hanyang University,\\
	Ansan 15588, Republic of Korea
}
\SetAuthorBlock{Hyogu Jeong}{%
	Narnia Labs,\\
	Daejeon 34051, Republic of Korea
}
\SetAuthorBlock{Namwoo Kang\CorrespondingAuthor}{%
	Associate Professor\\
	Cho Chun Shik Graduate School of Mobility,\\
	Korea Advanced Institute of Science and Technology, and Narnia Labs,\\
	Daejeon 34051, Republic of Korea\\
	e-mail: nwkang@kaist.ac.kr
}
\fi

\title{\ours{}: Foundation Model-Driven Large-Scale 3D Engineering Dataset via 2D Latent Space Augmentation}


\begin{abstract}
Data-driven engineering design is constrained by the lack of large-scale 3D datasets that pair geometry with physics-based performance labels. In particular, existing 3D data augmentation techniques have limitations in preserving subtle and diverse geometric variations, and it remains difficult to automate the subsequent simulation-labeling process, where boundary conditions vary depending on the generated geometry.
We present \ours{}, a foundation-model-driven data-augmentation framework that expands a small seed set of jet engine brackets into a large, simulation-labeled 3D dataset under constrained resources.
Our key idea is to augment in the data-rich 2D latent space, then transfer to 3D.
In Stage~1, we fine-tune a pretrained 2D latent diffusion model on multi-view renders and synthesize novel views by latent interpolation, retaining manufacturable designs through a vision-language-model (VLM) quality filter.
In Stage~2, the validated images are lifted to 3D meshes by a domain-adapted generative foundation model.
In Stage~3, an automated pipeline recognizes the load and bolt interfaces on each mesh and assigns finite-element labels---mass, stress, and displacement---without manual intervention.
We assess augmentation quality along three intrinsic axes: manufacturability, label fidelity against the SimJEB ground truth, and distributional consistency.
Starting from fewer than 400 seed designs, \ours{} yields 15{,}360 simulation-labeled 3D brackets---a $40\times$ expansion---using a single GPU per stage.
The dataset will be made publicly available to support reproducible engineering-AI research.
\end{abstract}

\maketitle
\ifblind\else\revfootnote{\textsuperscript{*}S. Yoo and L. Jeong contributed equally to this work.}\fi

\section{Introduction}
\label{sec:intro}

Data-driven methods are reshaping mechanical design, from generative design synthesis~\citep{oh2019deep,regenwetter2022deep,nie2021topologygan} to surrogate modeling of structural performance~\citep{cunningham2019investigation,pfaff2021meshgraphnets}.
The effectiveness of these methods scales with the size and quality of the datasets that support them.
Yet, unlike the natural-image domain---where billion-scale corpora are readily available---progress in engineering design is fundamentally constrained by the scarcity of large 3D datasets that pair geometry with physics-based performance labels.
Such data demand expert CAD modeling and high-fidelity simulation, both of which are costly and difficult to scale.

This bottleneck is evident even in the most established benchmarks.
The Simulated Jet Engine Bracket dataset (SimJEB)~\citep{whalen2021simjeb}, one of the few public 3D engineering datasets, provides 381 hand-designed brackets with finite element analysis (FEA) results.
DeepJEB~\citep{hong2025deepjeb} subsequently expanded this resource to 2{,}138 samples using a DeepSDF-based~\citep{park2019deepsdf} generative model with an automated simulation pipeline.
While these efforts mark important progress, their scale remains an order of magnitude below what modern deep architectures typically require, and each relies on a domain-specific generative model that is trained from scratch and must be rebuilt for every new domain, without access to the web-scale priors that now drive progress elsewhere.
Two further limitations stem from how DeepJEB augments and labels its data. Because it augments \emph{within} an implicit signed-distance field, the procedure tends to over-smooth the geometry and wash out the fine, locally varying features that distinguish individual brackets, narrowing the geometric diversity of the synthesized set relative to the original SimJEB designs. Moreover, to keep its labeling pipeline fully automatic, DeepJEB imposes a single, fixed boundary-condition template on every sample; this simplifies simulation but precludes a dataset that spans diverse boundary conditions.

In parallel, foundation models pretrained on web-scale data have begun to transfer well to engineering tasks, suggesting a path beyond hand-curated datasets.
Harnessing this capability, however, has so far demanded enormous computational resources and access to proprietary weights~\citep{physicsx2024gfm}, placing it out of reach for most academic and small-enterprise settings.
This motivates a central question for democratizing data-driven design: \emph{how can a small seed set be augmented into a large, simulation-labeled 3D dataset by adapting pretrained foundation models under severely limited resources?}
We address this question with \ours{}, a large-scale jet engine bracket dataset built by adapting pretrained 2D and 3D foundation models under tightly constrained resources.
Our key observation is that the asymmetry between 2D and 3D foundation models can be turned into an advantage: 2D generative models inherit priors from billions of images, whereas 3D models are trained on far smaller corpora.
This leads to our central idea: rather than augmenting directly in the original 3D space---where generative priors are scarce---we perform the augmentation in the data-rich 2D image space and then lift the results back to 3D.
\ours{} realizes this idea by exploiting the 2D--3D asymmetry through a three-stage pipeline.
Stage~1 augments a small seed set in the latent space of a fine-tuned 2D diffusion model and filters the synthesized images with a vision-language-model (VLM)-based quality classifier; in developing this classifier we identify and remedy a systematic negation artifact---which we term the \emph{Negative Words Negation} (NWN) problem---that arises when VLMs describe the \emph{absence} of defects.
Stage~2 lifts the validated images to 3D through domain-specific fine-tuning of a state-of-the-art 3D generative foundation model, which we show reconstructs reliably from even a single well-chosen viewpoint.
Stage~3 closes the loop with an automated computer-aided engineering (CAE) pipeline that detects load and bolt interfaces on the generated meshes and assigns physics-based structural performance labels without manual intervention.
Starting from fewer than 400 seed brackets, this pipeline produces an order-of-magnitude larger labeled 3D dataset using only a single GPU per training stage---demonstrating that domain-specific foundation-model pipelines, long assumed to require large-scale compute, are within reach of academic labs and small enterprises.

\section{Related Work}
\label{sec:related}

\subsection{3D Engineering Datasets}

Public 3D datasets for engineering applications remain scarce compared to the computer vision domain.
ShapeNet~\citep{chang2015shapenet} and ABC~\citep{koch2019abc} provide large collections of general 3D shapes but lack engineering performance labels.
Domain-specific efforts include FRAMED~\citep{regenwetter2023framed} (4{,}500 parametric bike frames with FEM results), BIKED~\citep{regenwetter2022biked} (4{,}500 bicycle designs with machine-learning benchmarks), Ship-D~\citep{bagazinski2023shipd} (ship hulls with drag coefficients), DrivAerNet++~\citep{elrefaie2024drivaernetpp} (car aerodynamics with CFD), and AircraftVerse~\citep{cobb2023aircraftverse} (27{,}714 multimodal aerial-vehicle designs).

For structural analysis, SimJEB~\citep{whalen2021simjeb} provides 381 jet engine brackets with FEA data, and DeepJEB~\citep{hong2025deepjeb} expanded this to 2{,}138 samples using a DeepSDF auto-decoder with automated simulation.
However, both remain an order of magnitude smaller than modern architectures demand, and---more fundamentally---DeepJEB augments \emph{within} a domain-specific implicit representation: a DeepSDF auto-decoder trained from scratch on the brackets themselves, so the augmentation inherits no prior beyond the seed set and each new domain requires retraining a bespoke generator.
Beyond scale, this implicit augmentation sacrifices two kinds of diversity present in the original SimJEB designs---the fine geometric variation across hand-designed brackets and the per-design variation in load and bolt interfaces---and recovering both was the starting motivation for \ours{}.
\ours{} takes a different route. Rather than training a domain-specific generator, it \emph{adapts} 2D and 3D foundation models pretrained on billions of images and hundreds of thousands of shapes, inheriting their external priors to expand the labeled dataset by an order of magnitude at single-GPU cost. It restores geometric diversity by augmenting in the data-rich 2D image space and recovers boundary-condition diversity by recognizing the load and bolt interfaces on each generated mesh, so that the constraints adapt to the individual geometry rather than following a single fixed template.
Because the pipeline reuses general-purpose foundation models rather than a bracket-specific implicit network, its augmentation backbone is in principle, domain-agnostic, with cross-domain transfer left to future work.

\subsection{Foundation Models for 3D Generation}

The landscape of 3D generative models has rapidly evolved with the adoption of foundation model paradigms.
Early image-to-3D systems such as Point-E~\citep{nichol2022pointe} and novel-view diffusion (Zero-1-to-3~\citep{liu2023zero123}), together with text-to-3D optimization (DreamFusion~\citep{poole2023dreamfusion}), established conditional 3D generation, which recent foundation models have sharply improved in fidelity.
TRELLIS~\citep{xiang2025trellis} introduces Structured 3D Latents (SLAT), a unified sparse latent representation that supports decoding into multiple output formats (NeRF~\citep{mildenhall2020nerf}, 3D Gaussians~\citep{kerbl2023gaussian}, meshes) via rectified flow~\citep{liu2023rectifiedflow} transformers.
Trained on over 500K 3D assets with up to 2 billion parameters, TRELLIS represents the current state-of-the-art in versatile 3D generation.

A parallel line of work targets fast feed-forward single-image-to-3D reconstruction: LRM~\citep{hong2024lrm} regresses a NeRF directly from one image, One-2-3-45~\citep{liu2023one2345} lifts multi-view diffusion outputs to a mesh in seconds, Wonder3D~\citep{long2024wonder3d} couples cross-domain (image and normal) diffusion for reconstruction, and InstantMesh~\citep{xu2024instantmesh} pairs a multi-view diffusion model with a sparse-view LRM.
Other notable models include TripoSG~\citep{li2025triposg} (SDF-based with rectified flow) and Hunyuan3D~\citep{hunyuan3d} (high-quality but restricted licensing).
We select TRELLIS for its MIT License, strong multi-view conditioning capabilities, and flexible output format---critical requirements for engineering applications where mesh quality directly impacts downstream simulation.

\subsection{Data Augmentation via Generative Models}

Generative data augmentation has been explored extensively in 2D~\citep{chen2021padgan,nobari2021pcdgan} and increasingly in 3D~\citep{shu20203d,wang2022ihgan}; diffusion models in particular have advanced performance-aware engineering design generation, such as for topology optimization~\citep{maze2023diffusion}.
However, most approaches operate within a single representation domain.
Our work is distinguished by its \textbf{cross-dimensional transfer strategy}: we augment data in 2D latent space (where billion-scale pretrained knowledge exists) and then reconstruct to 3D (where domain-specific fine-tuning ensures geometric fidelity).
This approach maximally leverages the knowledge embedded in large pretrained models while maintaining engineering validity through quality filtering. A data-augmentation method is judged by whether its samples are valid (on the data manifold), distributionally consistent with the seed set, and label-accurate; we assess \ours{} against these criteria in Section~\ref{sec:experiments}.

\subsection{Vision-Language Models for Quality Assessment}

Building on contrastive vision--language pretraining~\citep{radford2021clip}, VLMs such as BLIP-2~\citep{li2023blip2} and LLaVA~\citep{liu2023llava} have demonstrated strong visual understanding capabilities.
We repurpose these models for a novel task: automated engineering quality assessment of synthesized images.
Our approach differs from standard VLM applications in that it requires fine-grained detection of manufacturing defects (surface roughness, structural discontinuities) rather than high-level semantic understanding, motivating our choice of LLaVA's spatial information-preserving architecture over BLIP-2's text-centric compression.

\section{Method}
\label{sec:method}

\begin{figure*}[!htbp]
    \centering
    \includegraphics[width=0.95\textwidth]{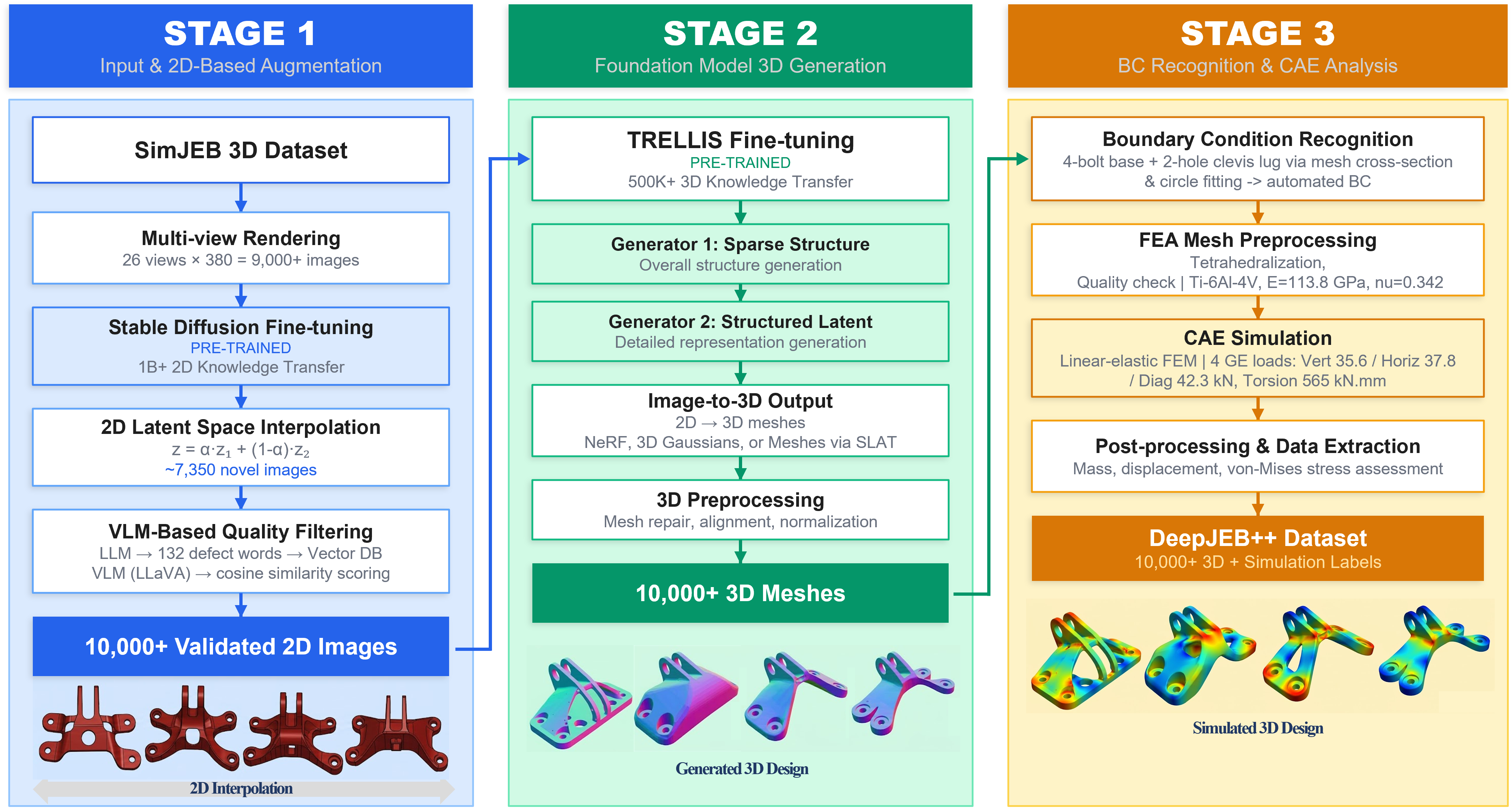}
    \caption{Overview of the \ours{} pipeline. Stage~1 augments 380 seed brackets into ${\sim}147{,}000$ candidate 2D images via Stable Diffusion fine-tuning and latent-space interpolation, which VLM-based quality filtering reduces to 22{,}495 validated designs. Stage~2 lifts these to 3D meshes using a domain-adapted TRELLIS model. Stage~3 performs automated boundary-condition recognition and CAE simulation, producing the final \ours{} dataset of 15{,}360 simulation-labeled 3D geometries.}
    \label{fig:pipeline}
\end{figure*}

Figure~\ref{fig:pipeline} illustrates the overall \ours{} pipeline, which consists of three main stages: 2D-based data augmentation (Stage~1), 3D mesh generation via foundation model fine-tuning (Stage~2), and boundary condition recognition with CAE analysis (Stage~3).

\subsection{Domain Dataset: SimJEB}
\label{sec:simjeb}

We use the SimJEB dataset~\citep{whalen2021simjeb} as our seed data source.
SimJEB consists of 381 hand-designed jet engine bracket CAD models from the GE Jet Engine Bracket Challenge, with associated FEA simulation results.
Following the data curation procedure of DeepJEB~\citep{hong2025deepjeb}, we retain 380 geometrically stable samples with consistent boundary conditions as our seed set.
Each bracket is characterized by a loaded interface and four bolted interfaces, imposing specific geometric constraints that generated designs must respect.

\subsection{Stage 1: 2D-Based 3D Data Augmentation}
\label{sec:stage1}

The core insight of Stage~1 is that 2D image generation models, pretrained on billions of images, encode rich geometric priors that can be transferred to engineering domains through efficient fine-tuning.
By operating in 2D latent space rather than directly in 3D, we can leverage this massive pretrained knowledge while sidestepping the data scarcity problem in 3D.

\subsubsection{Multi-View Rendering}
\label{sec:rendering}

Each of the 380 SimJEB bracket meshes is rendered from 26 viewpoints (8 azimuth angles $\times$ 3 elevation angles + top/bottom views), producing a total of 9{,}880 multi-view images.
Camera parameters (direction, focal distance, background type) are recorded in JSON format for each rendering, enabling precise view-conditioned generation in Stage~2.
We standardize the rendering conditions (lighting, resolution, background) to ensure consistency across all samples.

\subsubsection{Stable Diffusion Fine-Tuning}
\label{sec:sd_finetuning}

We fine-tune Stable Diffusion~\citep{rombach2022stablediffusion}, a latent diffusion model (building on denoising diffusion probabilistic models~\citep{ho2020ddpm}) pretrained on over 1 billion image-text pairs, on 7{,}800 SimJEB multi-view images (a 300-bracket training subset of the 380, rendered from 26 viewpoints each) using full parameter tuning.
The text prompt ``\texttt{A mechanical bracket design.}'' is used as the conditioning signal.
Training is conducted on a single A100 GPU for 1--2 days, yielding a model that generates bracket-specific geometries while retaining the general shape priors from pretraining.

This fine-tuning strategy is critical: without it, the model generates generic mechanical parts that lack the specific topological features of jet engine brackets.
With fine-tuning, the model consistently produces designs that respect the characteristic multi-interface structure of the bracket domain.

\subsubsection{Latent Space Interpolation}
\label{sec:interpolation}

\begin{figure*}[tp]
    \centering
    \includegraphics[width=\textwidth]{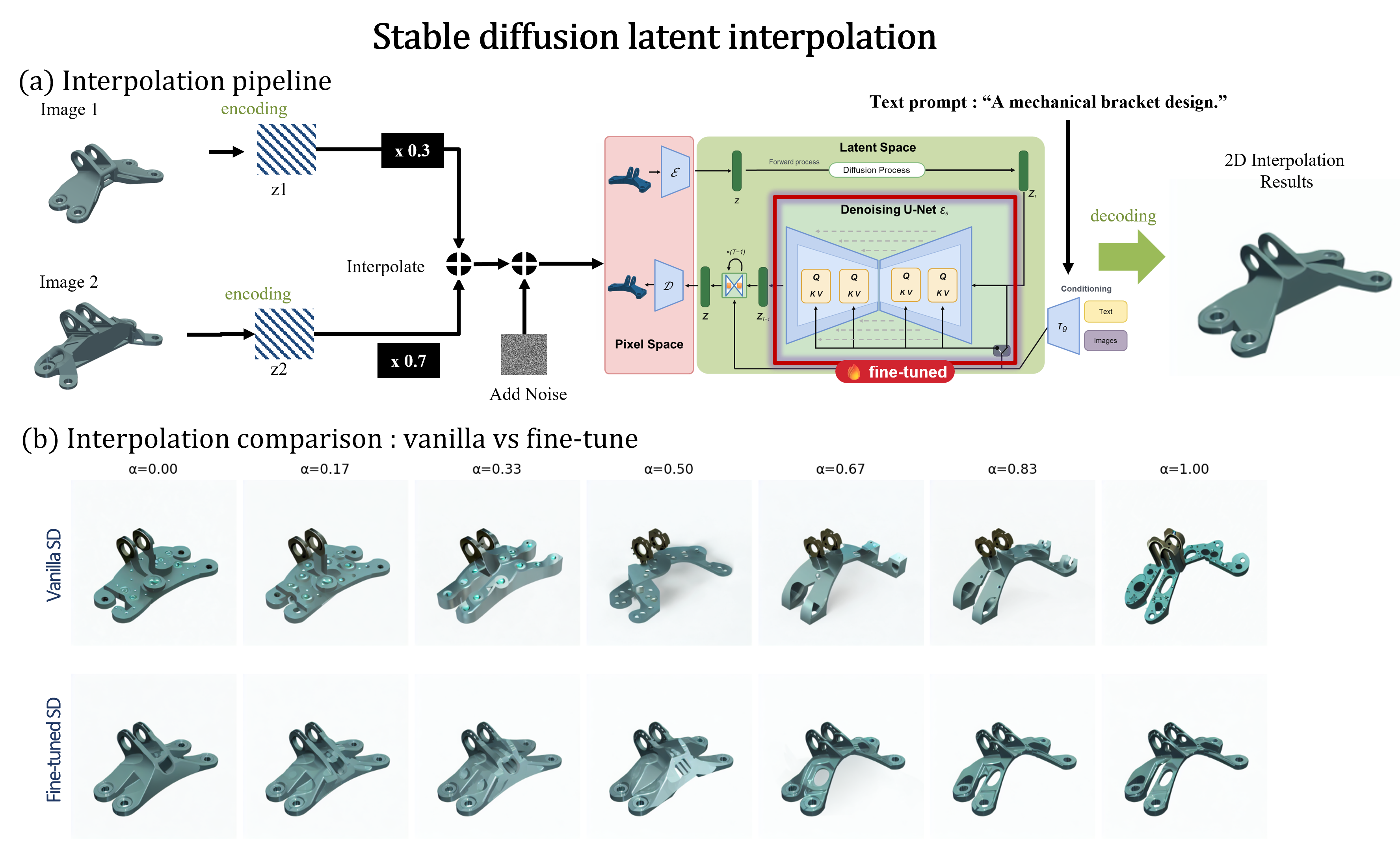}
    \caption{Stage-1 latent-space interpolation and its dependence on domain fine-tuning (Section~\ref{sec:interpolation}). \emph{(a)~Interpolation pipeline}: two seed bracket images are encoded, latent-blended, and decoded by the fine-tuned diffusion model. \emph{(b)~Interpolation comparison}: one interpolation path ($\alpha=0\!\rightarrow\!1$) decoded by the vanilla Stable Diffusion~v1.5 model (\emph{Vanilla SD} row) and our SimJEB-fine-tuned model (\emph{Fine-tuned SD} row).}
    \label{fig:sd_interp}
\end{figure*}

We generate novel bracket designs by interpolating between pairs of images~\citep{wang2023interpolating} in the latent space of the fine-tuned Stable Diffusion model.
Given two bracket images $I_1$ and $I_2$, we encode them into latent vectors $\mathbf{z}_1$ and $\mathbf{z}_2$, and compute:
\begin{equation}
    \mathbf{z}_{\text{interp}} = (1 - \alpha)\,\mathbf{z}_1 + \alpha\,\mathbf{z}_2
    \label{eq:interpolation}
\end{equation}
where the mix ratio $\alpha$ is swept from 0 to 1 over 19 evenly spaced steps (IS00--IS18, including the two endpoints) along each interpolation path.
After adding noise, the interpolated latent is decoded by the fine-tuned model with the text prompt condition; Figure~\ref{fig:sd_interp}(a) summarizes this encode--blend--noise--decode pipeline.

\paragraph{Representative Pair Selection.}
Na\"ive random pairing would produce many redundant interpolations between geometrically similar brackets.
To ensure diversity, we apply the following selection procedure:
\begin{enumerate}
    \item Convert all 380 bracket images to grayscale and normalize pixel intensities.
    \item Apply Principal Component Analysis (PCA) for dimensionality reduction.
    \item Score each sample by Euclidean distance in the reduced space.
    \item Select 50 representative samples via 1D equal allocation across the distance distribution.
\end{enumerate}
This yields $\binom{50}{2} = 1{,}225$ possible image pairs, of which $1{,}184$ are generated; interpolating each pair across the mix-ratio steps above, with 8 rendered views per step, yields a pool of $\approx\mathbf{147{,}000}$ candidate 2D images before quality filtering.

\paragraph{Effect of Fine-Tuning on Interpolation.}
Domain fine-tuning is essential for the interpolation to stay within the bracket domain.
Figure~\ref{fig:sd_interp}(b) compares interpolations between the same two SimJEB brackets---shown as seven evenly spaced frames of the 19-step path---decoded by the vanilla Stable Diffusion v1.5 model and by our SimJEB-fine-tuned model.
The fine-tuned model yields valid, smoothly varying bracket geometries along the entire interpolation path, whereas the vanilla model collapses to generic mechanical parts (gear- and turbine-like structures) that violate the multi-interface bracket topology.
Each panel---including the endpoints ($\alpha\!=\!0,1$)---is re-synthesized by the model from the interpolated latent through image-to-image generation, so every panel reflects the model's own generative prior rather than a direct copy of the input; consequently the vanilla model distorts even the endpoints.
This is not an encoding artifact: a pure VAE encode--decode round-trip of the endpoints is near-lossless, so the distortion stems from the bracket-unaware vanilla prior, not from the latent embedding.

\subsubsection{VLM-Based Quality Classifier}
\label{sec:vlm_classifier}

Not all synthesized images represent valid, manufacturable bracket designs.
We develop an automated quality filtering framework using Vision-Language Models to assess each candidate image against engineering criteria.

\begin{figure*}[tp]
    \centering
    \includegraphics[width=0.95\textwidth]{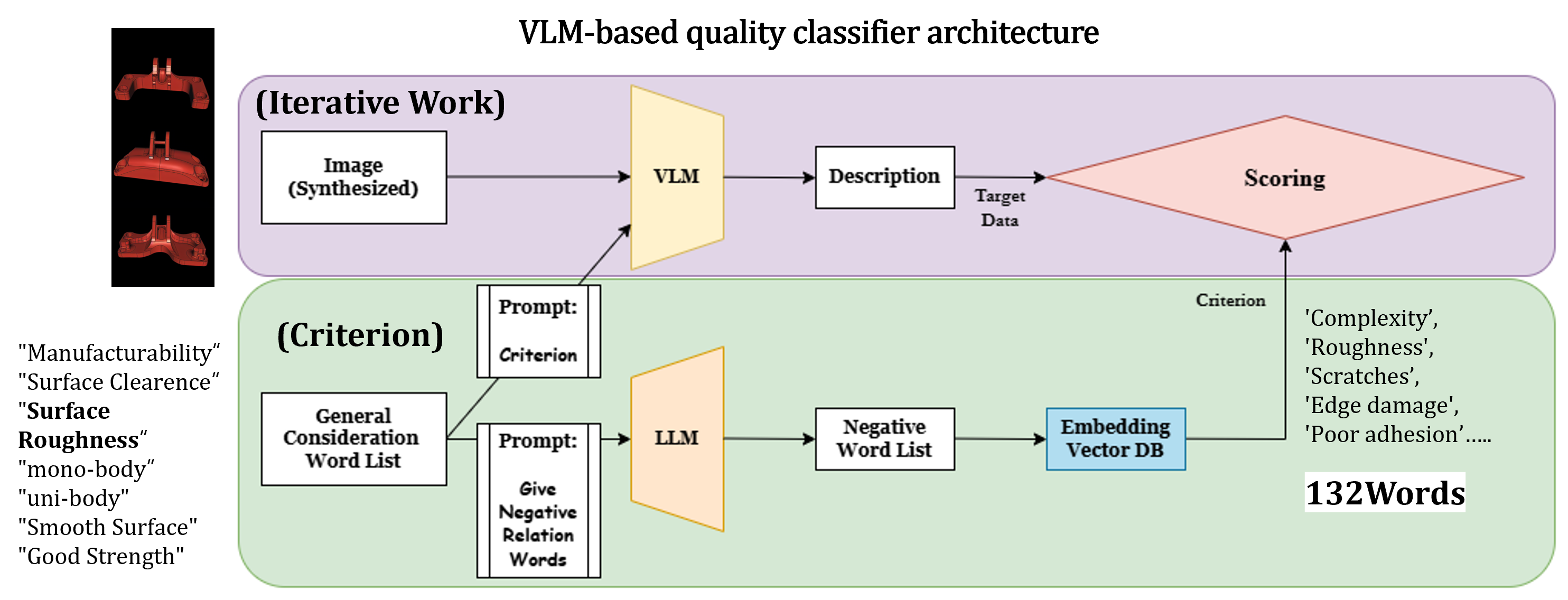}
    \caption{VLM-based quality classifier architecture. The LLM generates domain-specific negative word embeddings, while the VLM produces image descriptions. Quality is assessed by the semantic distance between descriptions and defect vocabulary.}
    \label{fig:vlm}
\end{figure*}

\paragraph{Architecture.}
The classifier consists of three components (Figure~\ref{fig:vlm}):
\begin{enumerate}
    \item \textbf{LLM branch:} Given engineering criteria (manufacturability, surface clearance, surface roughness, mono-body structure, smooth surface, good strength), a large language model generates a set of 132 \textit{negative words} describing potential defects (e.g., ``scratches,'' ``cracks,'' ``poor adhesion'').
    These words are embedded into a vector database.
    
    \item \textbf{VLM branch:} Each synthesized image is processed by a VLM with a prompt requesting description of the surface and structural characteristics. The VLM outputs a textual description of the image.
    
    \item \textbf{Scoring:} The cosine similarity between the VLM description embedding and the negative word embeddings is computed; higher similarity indicates more defects. An image is filtered out when its score lies in the most defect-like top-$p\%$, with the threshold $p$ calibrated on a labeled benchmark (Section~\ref{sec:exp_stage1}).
\end{enumerate}

\paragraph{Negative Words Negation (NWN) Problem.}
We identify a subtle failure mode in VLM-based quality assessment.
When a VLM describes a high-quality image, it may use phrases like ``no scratches'' or ``not rough.''
These descriptions contain negative words that match the defect vocabulary, causing the similarity score to incorrectly penalize high-quality images.
We term this the \textit{Negative Words Negation} (NWN) problem.

We resolve this by adding an explicit constraint to the VLM prompt: ``\texttt{Do not use the word `no'.}''
This forces the VLM to use affirmative descriptions (e.g., ``the surface is smooth'' instead of ``no scratches''), eliminating the double-negation artifact.
This simple intervention improves classifier accuracy from 56.45\% to \textbf{72.17\%} on the labeled benchmark---the 636 decisive images among 900 manually labeled samples (the borderline ``acceptably corrupt'' class is excluded; Section~\ref{sec:exp_stage1}).

\paragraph{VLM Model Selection.}
We compare two VLM architectures for the quality assessment task:
\begin{itemize}
    \item \textbf{BLIP-2}~\citep{li2023blip2}: Uses a Q-Former for text-centric information compression. The heavy compression step risks losing fine-grained local features (surface roughness, micro-cracks) that are critical for engineering quality assessment.
    \item \textbf{LLaVA}~\citep{liu2023llava}: Uses a projection matrix to map visual features directly to the language model, preserving patch-level spatial information. This architecture retains local defect information more effectively.
\end{itemize}
We select LLaVA for our pipeline due to its superior preservation of local geometric features, which is essential for detecting surface-level manufacturing defects.

The VLM filter is applied per candidate image; from the validated frames we then lift the diagonal view of each interpolation step across the $1{,}184$ retained pairs to 3D, yielding \textbf{22{,}495 \ours{} designs} from the ${\sim}147{,}000$-image candidate pool.

\subsection{Stage 2: 3D Generation via Foundation Model Fine-Tuning}
\label{sec:stage2}

\subsubsection{Model Selection}
\label{sec:model_selection}

We evaluate three candidate image-to-3D foundation models: Hunyuan3D~2.0~\citep{hunyuan3d}, TripoSG~\citep{li2025triposg}, and TRELLIS~\citep{xiang2025trellis}.
Licensing is a decisive criterion, since we intend to release the dataset openly and the generated geometries inherit the generator's license: Hunyuan3D~2.0 carries regional usage restrictions (e.g., EU and Korea), and TripoSG relies on restrictively licensed dependencies, whereas TRELLIS is distributed under the permissive MIT License.
We therefore select TRELLIS, which additionally supports multi-view conditioning and decodes its structured latents (SLAT) into multiple 3D formats (NeRF, 3D Gaussians, and meshes).

\subsubsection{TRELLIS Architecture}
\label{sec:trellis_arch}

TRELLIS employs a two-generator architecture operating on Structured 3D Latents (SLAT):
\begin{itemize}
    \item \textbf{Sparse Structure Generator (SS-Generator):} Predicts which voxels in a 3D grid are active (i.e., where geometry exists), conditioned on input images. Uses a rectified flow transformer.
    \item \textbf{Structured Latent Generator (SLAT-Generator):} Fills the active voxels with detailed shape and appearance features, also based on a sparse flow transformer architecture.
\end{itemize}
This decomposition enables efficient generation by operating only on sparse, occupied regions of 3D space, substantially reducing computational cost compared to dense volumetric approaches.

\subsubsection{Domain-Specific Fine-Tuning}
\label{sec:trellis_finetuning}

We fine-tune both generators of TRELLIS on the SimJEB dataset to adapt the pretrained model (originally trained on general 3D assets) to the engineering bracket domain.

\paragraph{Data Preparation Pipeline.}
For each of the 380 SimJEB meshes:
\begin{enumerate}
    \item Render 150 multi-view images and extract DINO~\citep{caron2021dino} visual features for each view.
    \item Voxelize the 3D mesh to obtain sparse structure encoding.
    \item Combine DINO features with sparse structure encoding to produce SLAT encoding.
\end{enumerate}
The full data preparation requires approximately 10 hours.
We use 300 samples for training and 80 for testing.

\paragraph{Training Protocol.}
\begin{itemize}
    \item \textbf{Generator 1 (SS-Generator):} Fine-tuned on 25 multi-view images per sample (a subset of the 150 rendered views) with corresponding sparse structure labels.
    \item \textbf{Generator 2 (SLAT-Generator):} Fine-tuned with single-image conditioning to generate detailed 3D representations from the structure predicted by Generator~1.
\end{itemize}

\subsubsection{Inference and 3D Preprocessing}
\label{sec:inference}

The fine-tuned TRELLIS model converts validated 2D bracket images from Stage~1 into 3D meshes via SLAT decoding into NeRF, 3D Gaussians, or triangle meshes.
The model supports both single-view and multi-view inference:
\begin{itemize}
    \item \textbf{Single-view:} Each 2D synthesized image directly produces a 3D mesh.
    \item \textbf{Multi-view:} When multiple views of the same object are available, combining them improves reconstruction accuracy, saturating by about six views (Section~\ref{sec:exp_stage2}); a single informative (diagonal) viewpoint already approaches this level.
\end{itemize}

After 3D generation, a preprocessing step performs mesh repair, alignment, and normalization to ensure geometric consistency across the generated dataset.
This step corrects common artifacts from the image-to-3D conversion, such as non-manifold edges, self-intersections, and inconsistent orientation, producing clean triangle meshes suitable for downstream FEA preprocessing in Stage~3.

\subsection{Stage 3: Boundary Condition Recognition and CAE Analysis}
\label{sec:stage3}

While Stages 1 and 2 produce 3D bracket meshes, engineering datasets require associated simulation labels for training surrogate models and evaluating structural performance. Stage 3 closes this gap through a fully automated pipeline that recognizes the load and bolt interfaces on each generated mesh, assigns physically consistent boundary conditions, and carries out Finite Element Analysis (FEA) to extract performance labels.

\subsubsection{Automated Interface Detection}
\label{sec:bc_recognition}


Bolt interfaces are detected directly on the mesh. Lower-flange vertices with normals perpendicular to the bolt axis are grouped into connected components via mesh adjacency rather than Euclidean clustering, which keeps adjacent bores separate and a single bore intact across a thick flange. A component is classified as a cylindrical bore when its vertex normals sweep $\geq 270^\circ$ about the axis and face inward. Coaxial components are merged, retaining only the smallest-diameter ring and discarding the larger counterbores. Selecting the outermost bore in each corner of the flanges yields the four clamped bolt interfaces.


The loaded clevis interface is identified by sectioning the mesh with planes perpendicular to the pin axis. Each prong bore then appears as a closed loop, to which a circle is fitted via the K\aa sa method \citep{kasa1976circle}. The loop centers are clustered in 3D, and the pair of equal radius bores separated by $\geq 20$\,mm is taken as the load-carrying two-hole clevis. For generated meshes, the four detected bolt centers are rigidly registered to a fixed SimJEB bolt template via Umeyama's method \citep{umeyama1991}. The registration maps every mesh into the template's global coordinate frame, keeping the applied boundary conditions consistent with the original SimJEB problem definition.

\paragraph{Geometric Validation of Detected Lugs}

We validate each detected clevis with three pose-invariant geometric features computed from the fitted rings. The minimum angular coverage of the two bores, $\mathrm{cov}_{\min}$, distinguishes a closed bore from a partial arc. The perpendicular offset of the two ring centers from the common pin axis, $d_{XZ}$, measures coaxiality. Their spacing along that axis, $d_Y$, is the clevis gap. A closed bore covers nearly the full circle at about $354^\circ$, whereas a partial arc spans only about $257^\circ$, so a candidate is accepted when $\mathrm{cov}_{\min}\!\geq\!330^\circ$, $d_{XZ}\!\leq\!3\,\mathrm{mm}$, and $d_Y\!\geq\!18\,\mathrm{mm}$.


We apply this rule to the full 381-bracket SimJEB set, which is disjoint from the 380-bracket augmentation seed. A separate manual review of the accepted candidates serves as the reference. Across 332 candidates, the rule matches this review on all but one borderline case, an agreement of $99.7\%$, keeping 294 genuine clevises and rejecting 38 spurious detections without any learned model. These 294 brackets, 77\% of the 381, form the reference set for FEA validation.

\begin{figure*}[tp]
    \centering
    \includegraphics[width=\textwidth]{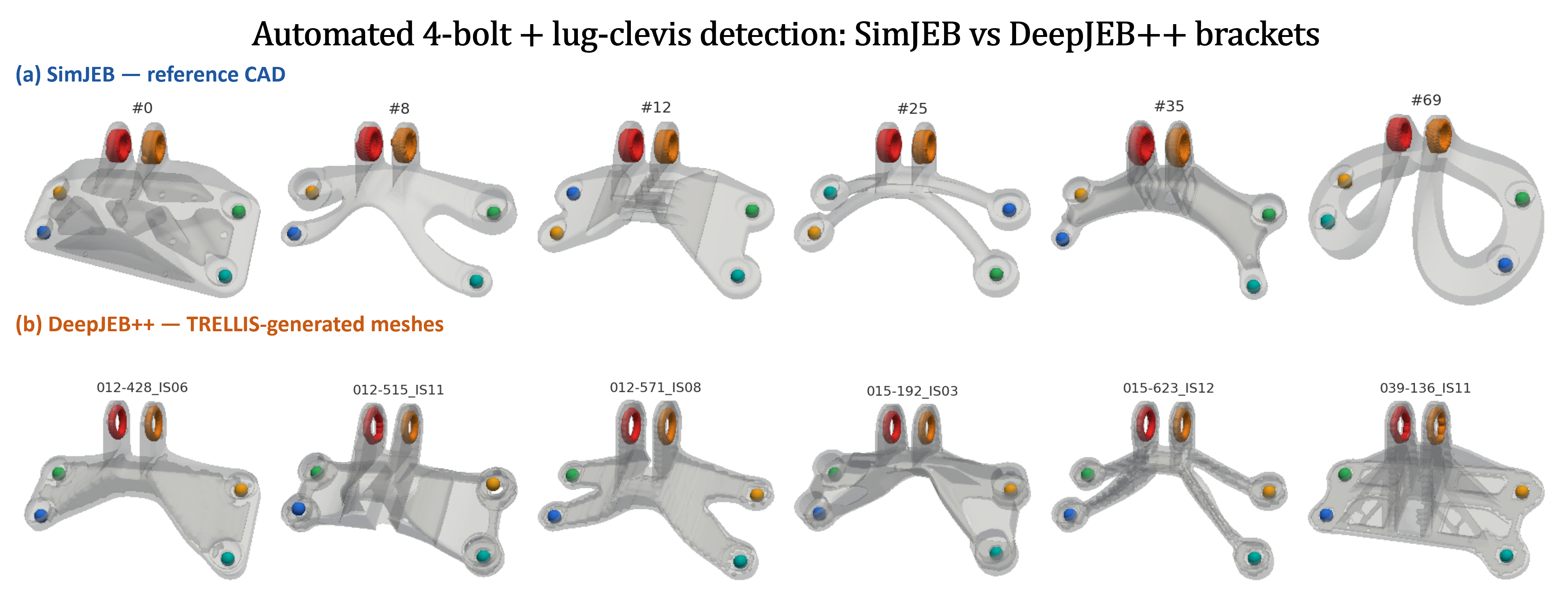}
    \caption{Automated four-bolt and lug--clevis interface detection applied to (a) real SimJEB reference CAD and (b) \ours{} TRELLIS-generated meshes. The same geometric detector recovers consistent interfaces on both real and generated geometry, enabling identical boundary-condition assignment. Detected bolt centers are shown as colored markers and the two clevis bores as rings.}
    \label{fig:det_compare}
\end{figure*}

Figure \ref{fig:det_compare} shows the same detector applied to both real SimJEB CAD and generated \ours{} meshes, detecting consistent four-bolt and lug-clevis interfaces on both. Generated meshes have thinner and slightly less coaxial interfaces, so the acceptance criterion is recalibrated on human labels (187 cases, about $99\%$ agreement) and relaxed to $\mathrm{cov}_{\min}\!\geq\!300^\circ$, $d_{XZ}\!\leq\!6\,\mathrm{mm}$, and $R\!\in\![7,11.5]\,\mathrm{mm}$. Meshes that fail this criterion are flagged as invalid and excluded from FEA, so that only geometrically valid interfaces receive simulation labels.

\subsubsection{Linear-Elastic Finite Element Solver}
\label{sec:fea_solver}


Each surface mesh is decimated to about 25{,}000 nodes without altering its topology, repaired into a single watertight surface, and tetrahedralized. The mesh quality is controlled by a radius--edge ratio below $1.2$ and a minimum dihedral angle of $10^\circ$. Boundary-node ordering is preserved throughout, so that detected interface nodes retain their indices.


We solve linear elasticity, with bolt nodes fully clamped and the clevis interface subjected to a distributed nodal load, or a distributed nodal moment for the torsional case. Material properties and loads strictly follow the SimJEB and GE Jet Engine Bracket Challenge specification. The material is Ti--6Al--4V, with $E=113.8$\,GPa, $\nu=0.342$, and a yield stress of $903$\,MPa ($131$\,ksi at the service temperature, as specified by the GE Jet Engine Bracket Challenge). Four load cases are applied, a vertical load of $35.6$\,kN, a horizontal load of $37.8$\,kN, a diagonal load of $42.3$\,kN at $42^\circ$ from vertical, and a torsional load of $565$\,kN$\cdot$mm. Per element stresses are mapped to volume-weighted nodal von Mises values.  For each load case we record the maximum nodal displacement and the von Mises stress. We report the 95th-percentile von Mises stress, $\sigma_{\text{vm}}^{p95}$, as the primary stress label, since it is robust to the single-point stress singularities that sharp generated edges can produce and that otherwise dominate the raw maximum. Together with the 3D geometries, these per-load displacement and stress labels constitute the simulation-annotated \ours{} dataset.

\section{Experiments}
\label{sec:experiments}

\subsection{Stage 1: 2D Augmentation Quality}
\label{sec:exp_stage1}

\subsubsection{Stable Diffusion Fine-Tuning Effect}

Table~\ref{tab:sd_finetuning} summarizes the fine-tuning configuration and qualitative effect.
Without fine-tuning, Stable Diffusion generates generic mechanical parts.
After fine-tuning on SimJEB multi-view images, the model consistently produces bracket-specific topologies with appropriate interface structures.

\begin{table}[tbp]
\centering
\caption{Stage-1 configuration: Stable Diffusion fine-tuning and latent-space interpolation.}
\label{tab:sd_finetuning}
\resizebox{\columnwidth}{!}{%
\begin{tabular}{ll}
\toprule
\textbf{Component} & \textbf{Setting} \\
\midrule
\multicolumn{2}{l}{\textit{Backbone \& fine-tuning}}\\
\quad Base model        & Stable Diffusion v1.5 \\
\quad Adapted module     & Full U-Net (VAE, text encoder frozen) \\
\quad Training images     & 7{,}800 SimJEB multi-view renders ($512{\times}512$) \\
\quad Text prompt         & ``A mechanical bracket design'' \\
\quad Optimizer           & AdamW \\
\quad Learning rate       & $1\times10^{-5}$ (constant) \\
\quad Batch size          & 16 (grad.\ accum.\ 1), fp32 \\
\quad Training steps       & 18{,}500 \\
\quad Compute             & $1\times$ NVIDIA A100 (80\,GB), $\sim$1--2 days \\
\midrule
\multicolumn{2}{l}{\textit{Latent-space interpolation (image-to-image)}}\\
\quad Interpolant         & $\mathbf{z}_{\mathrm{mix}}=(1-\alpha)\mathbf{z}_1+\alpha\mathbf{z}_2$ (VAE latent) \\
\quad Weights $\alpha$    & $\{0,0.2,0.4,0.6,0.8,1.0\}$ \\
\quad Noise strength      & $0.8$ \\
\quad Sampler / steps     & UniPC~\citep{zhao2023unipc} multistep / $20$ \\
\quad Guidance scale      & $7.5$~\citep{ho2022cfg} \\
\quad Seeds per pair      & $20$ \\
\bottomrule
\end{tabular}}
\end{table}

\subsubsection{VLM Classifier Performance}

\begin{figure*}[tp]
    \centering
    \includegraphics[width=\textwidth]{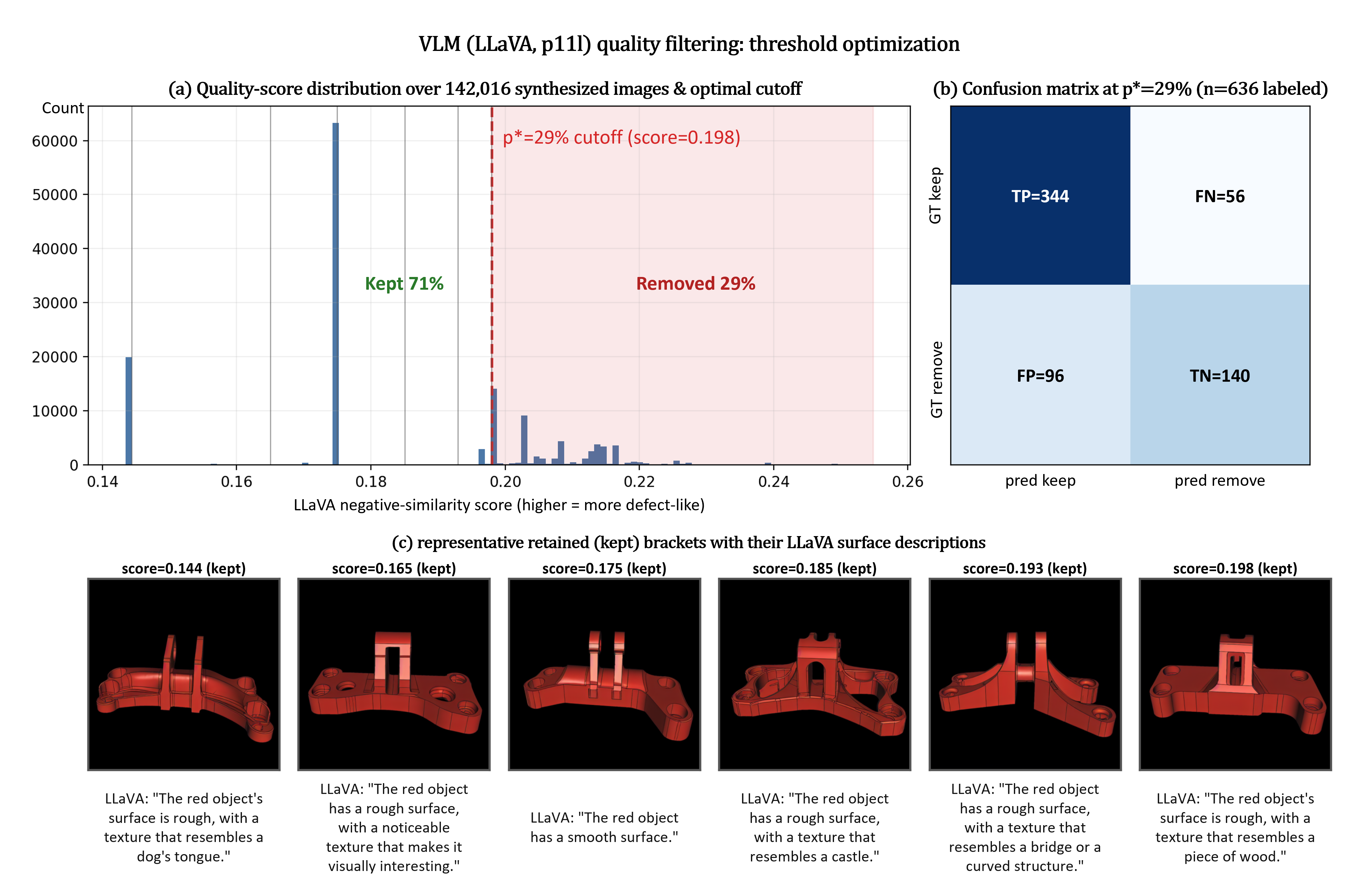}
    \caption{LLaVA-based quality filtering. (a) Distribution of the quality score (cosine similarity between the LLaVA surface description and the defect vocabulary) over all synthesized images, with the adopted top-$p{=}29\%$ cutoff separating kept from removed designs. (b) Confusion matrix on the 636-sample labeled benchmark at that operating point. (c) Representative retained brackets and their LLaVA surface descriptions; the score is a negative-word similarity rather than a literal roughness axis, so retained brackets may still be described as ``rough.''}
    \label{fig:vlm_results}
\end{figure*}

Each candidate image receives a quality score equal to the cosine similarity between its LLaVA-generated surface description and a defect (negative-word) vocabulary; a higher score indicates a more defect-like description.
Figure~\ref{fig:vlm_results} summarizes the filter: the score distribution over all synthesized images with the $p{=}29\%$ cutoff, the confusion matrix on the labeled benchmark at that operating point, and representative retained brackets together with their LLaVA descriptions.

To quantify filtering quality, we construct a benchmark of 900 synthesized images manually labeled as non-corrupt (400, to be kept), severely corrupt (236, to be removed), and acceptably corrupt (264).
The acceptably-corrupt class is borderline and subjective, so we exclude it from the accuracy computation, leaving 636 decisive samples.
The filter removes an image when its quality score lies in the highest-scoring (most defect-like) top-$p\%$ and keeps it otherwise; we adopt a conservative removal rate of $p=29\%$, discarding the most defect-like $29\%$ of images.
With the NWN-corrected prompt, accuracy on the labeled benchmark rises from 56.45\% (vanilla prompt) to 72.17\% at the default threshold, and at the adopted $p=29\%$ reaches \textbf{76.10\%} (false-positive rate 40.7\%, false-negative rate 14.0\%).
Because $p$ is a single fixed operating point rather than a per-sample-tuned quantity, the reported accuracy carries negligible selection optimism, which we confirm with two checks. First, on a larger labeled benchmark of $908$ images (an expansion of the $636$-sample set with additional frames), the filter attains $76.7\%$ accuracy at the same $p{=}29\%$, essentially reproducing the value above and indicating that the operating point is not specific to one benchmark. Second, even when $p$ is instead \emph{selected} to maximize benchmark accuracy, repeated five-fold cross-validation yields a held-out accuracy of $81.8\%$ against an in-sample optimum of $82.4\%$---a gap of only $0.6$ percentage points---so choosing the removal rate on a labeled set introduces no material optimism. The filter is moreover deliberately operated as a \emph{coarse first-stage screen}: its loose false-positive rate ($40.7\%$) discards only the most defect-like images, while final geometric and physical validity is enforced downstream by the Stage-3 interface-detection and FEA gates (Section~\ref{sec:stage3}). The augmentation therefore does not depend on the VLM filter being precise.

\subsubsection{Data Augmentation Summary}

Starting from 380 SimJEB samples and 26 views each (9{,}880 images), our pipeline produces:
\begin{itemize}
    \item 50 representative samples selected via PCA + Euclidean distance scoring
    \item 1{,}225 possible image pairs, of which 1{,}184 are generated
    \item $\sim$147{,}000 candidate 2D images (8 rendered views per interpolation step)
    \item \textbf{22{,}495 \ours{} designs} lifted to 3D in Stage~2 (diagonal view of each interpolation step, after VLM filtering)
\end{itemize}

\subsection{Stage 2: 3D Generation Quality}
\label{sec:exp_stage2}

\subsubsection{Single-View Reconstruction}

\begin{figure*}[tp]
    \centering
    \includegraphics[width=0.8\textwidth]{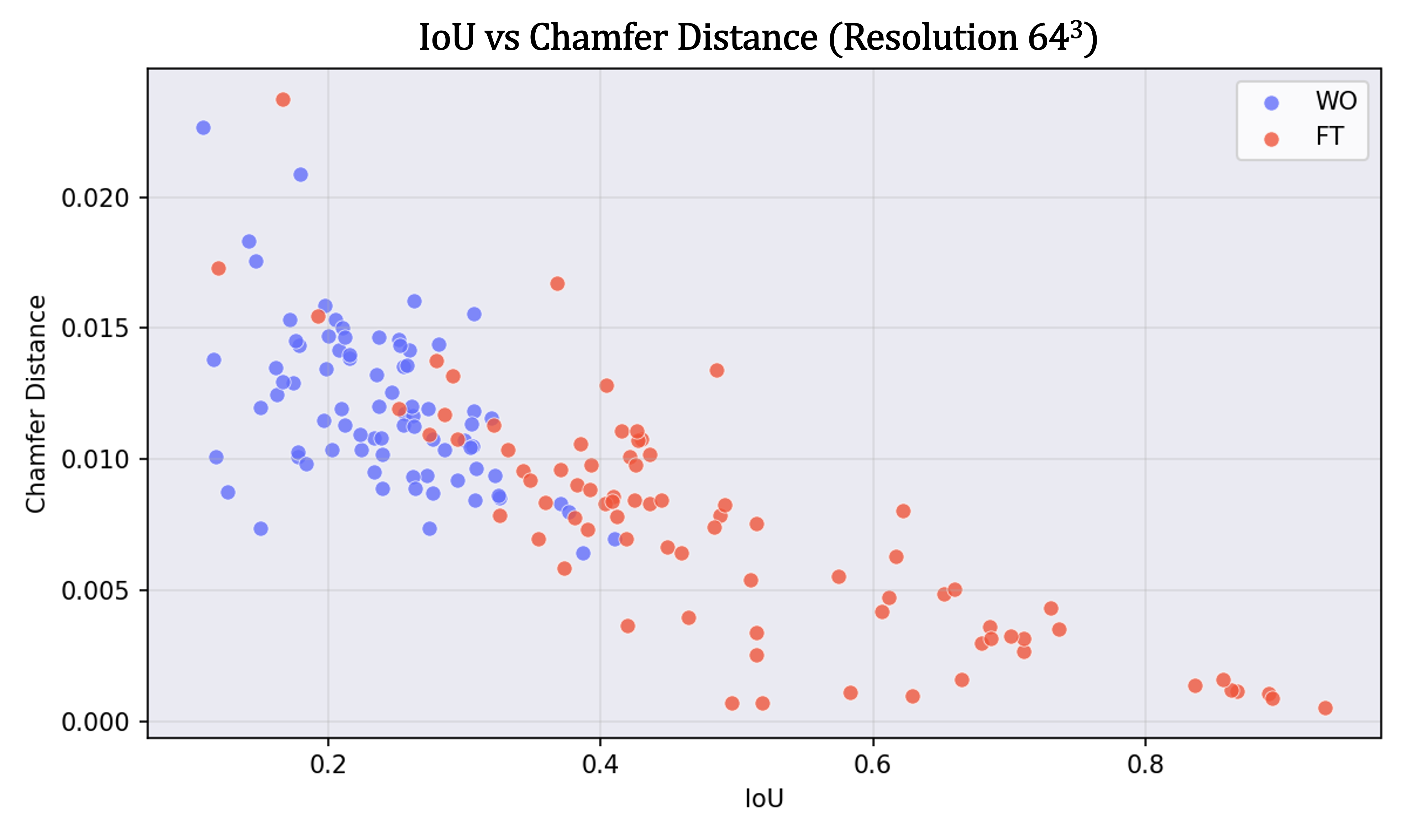}
    \caption{IoU vs.\ Chamfer Distance ($64^3$ resolution) for the 80 test brackets. Reconstructions by the fine-tuned model (red, FT) attain higher IoU and lower Chamfer Distance than the pretrained baseline (blue, WO) for the large majority of brackets (means in Table~\ref{tab:trellis_results}), with the two distributions overlapping only in the mid-quality region---demonstrating the benefit of domain-specific adaptation.}
    \label{fig:recon_scatter}
\end{figure*}

We evaluate the fine-tuned TRELLIS model on 80 held-out SimJEB test brackets using single-view inference.
Table~\ref{tab:trellis_results} shows that domain-specific fine-tuning substantially improves reconstruction quality, and Figure~\ref{fig:recon_scatter} confirms that this gain holds per bracket rather than only on average.

\begin{table*}[tp]
\centering
\caption{TRELLIS single-view inference results on SimJEB test set (80 samples).}
\label{tab:trellis_results}
\begin{tabular}{lccc}
\toprule
\textbf{Metric} & \textbf{No Fine-tuning} & \textbf{Fine-tuned} & \textbf{Improvement} \\
\midrule
IoU $\uparrow$ & 0.2390 & \textbf{0.4940} & +106.69\% \\
Chamfer Distance $\downarrow$ & 0.01199 & \textbf{0.00731} & $-$38.98\% \\
\bottomrule
\end{tabular}
\end{table*}

The pretrained TRELLIS model, trained on general 3D assets, fails to capture the specific structural patterns of engineering brackets (multi-interface topology, load-bearing arm structures).
Fine-tuning on only 300 SimJEB training samples is sufficient to recover these domain-specific features.

\subsubsection{Multi-View Conditioning}

\begin{figure*}[tp]
    \centering
    \includegraphics[width=\textwidth]{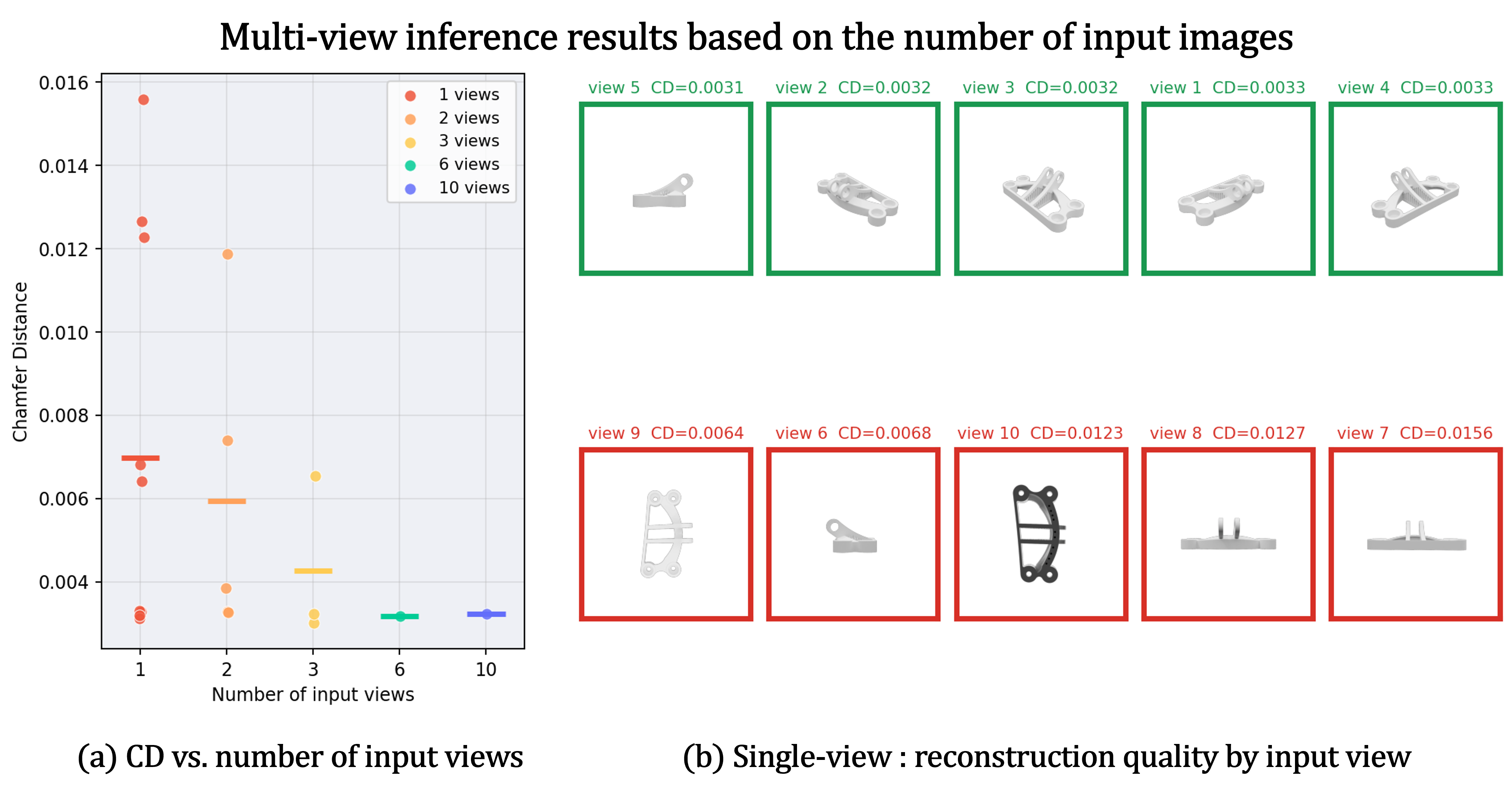}
    \caption{Reconstruction quality versus input views on a held-out SimJEB bracket. (a) Chamfer Distance for $\{1,2,3,6,10\}$ input views; the best single (diagonal) view already matches the six-view level. (b) Single-view reconstruction by viewpoint, sorted best to worst: diagonal three-quarter views (green) reconstruct well and are the viewpoint used for large-scale single-view generation, while top/front/bottom views (red) are degenerate.}
    \label{fig:multiview}
\end{figure*}

We investigate how reconstruction quality depends on the input views---both their number and their viewpoint---on a held-out SimJEB bracket.
Figure~\ref{fig:multiview} reports Chamfer Distance for $\{1,2,3,6,10\}$ input views (taken as contiguous chunks of the rendered viewpoints) and, for single-view inference, broken down by viewpoint.
Chamfer Distance decreases as more views are added and saturates by about six views.
Crucially, the \emph{best} single view---a diagonal three-quarter view---already attains a Chamfer Distance ($0.0031$) essentially equal to the six-view result ($0.0032$), whereas degenerate viewpoints (top, front, bottom) are far worse ($0.006$--$0.016$).
View \emph{type} therefore matters more than view \emph{count}: a single informative viewpoint suffices, and additional views mainly hedge against poor viewpoints.

This has a direct practical implication for our pipeline. Because the 2D designs synthesized by latent interpolation in Stage~1 have no pre-existing 3D object to re-photograph, multi-view conditioning is not available and large-scale generation necessarily relies on single-view inference. A well-chosen diagonal viewpoint matches the six-view reconstruction fidelity, so this choice incurs little loss; the fixed diagonal view used for generation is precisely the best-performing single viewpoint here.

\subsubsection{2D Synthetic Image to 3D Generation}

\begin{figure*}[tp]
    \centering
    \includegraphics[width=\textwidth]{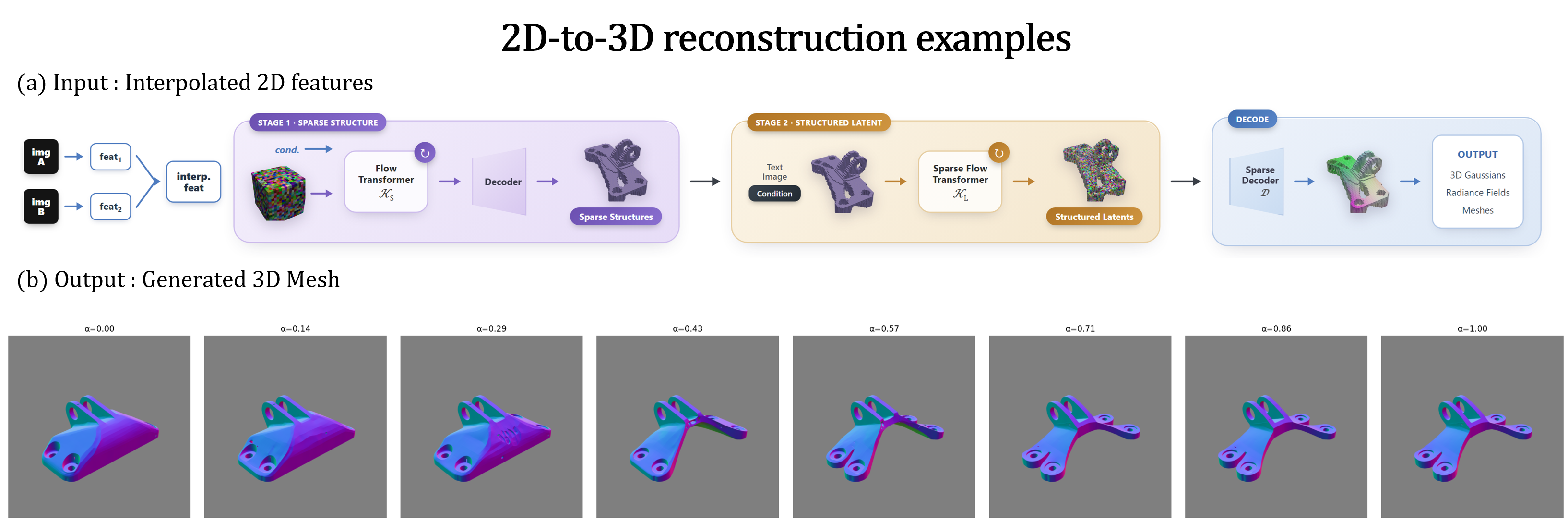}
    \caption{End-to-end 2D-to-3D generation in \ours{} (Section~\ref{sec:stage2}). \emph{(a)~Input: interpolated 2D features} ($\mathrm{img}\,A$, $\mathrm{img}\,B$ encoded and blended into \emph{interp.\ feat}) conditioning the domain-adapted TRELLIS model. \emph{(b)~Output: generated 3D mesh} along one interpolation path ($\alpha=0\!\rightarrow\!1$).}
    \label{fig:2d_to_3d}
\end{figure*}

We demonstrate that the 2D synthesized images from Stage~1 can be successfully converted to 3D meshes by the fine-tuned TRELLIS model.
Figure~\ref{fig:2d_to_3d}(a) shows the end-to-end architecture: the interpolated 2D features condition the domain-adapted TRELLIS model, which generates in two flow stages---sparse-structure generation followed by structured-latent generation (Section~\ref{sec:trellis_arch})---and decodes the structured latents into the output 3D representation (3D Gaussians, radiance fields, or meshes). Figure~\ref{fig:2d_to_3d}(b) shows the resulting meshes along one interpolation path (eight evenly spaced frames), illustrating that latent space interpolation in 2D translates to smooth, continuous shape variation in 3D.

\subsection{Stage 3: Solver and Mass-Label Validation against SimJEB Ground Truth}
\label{sec:exp_stage3}


Our dataset labels are the FEA results computed on the generated meshes themselves, not a reconstruction of any particular reference design. We accordingly validate the labeling pipeline, both its finite-element solver and its mass computation, on the real SimJEB meshes, where ground-truth values are available.


All 381 SimJEB CAD brackets carry ground-truth FEA labels. Here, 294 pass automated interface detection (Section~\ref{sec:bc_recognition}), 289 yield a stable node-level fit, and 279 are single-body meshes with valid mass labels, which serve as the comparison reference (Section~\ref{sec:sj_compare}). The 380-bracket augmentation seed (Section~\ref{sec:simjeb}) is the geometrically valid subset of the 381.

To quantify the pipeline's accuracy on a given mesh, we compare its output against the original SimJEB FEA results on the original brackets with validated interface detection (294 of 381 cases; Section~\ref{sec:bc_recognition}), under the identical material and loading configurations.

A direct comparison of scalar peaks is confounded: the applied load direction is recovered from geometry, and the decimated meshes differ from the reference discretization, so peak stresses and displacements mix load-direction uncertainty with solver fidelity.
We therefore isolate solver fidelity with a node-level test.
Because the elastic response is linear (by superposition) in the interface load, we solve six unit responses at the loaded interface---three forces ($F_x,F_y,F_z$) and three moments ($M_x,M_y,M_z$)---and least-squares fit their six coefficients to the SimJEB ground-truth nodal displacement field.
The resulting coefficient of determination $R^2$ measures how closely the two solvers agree once load-direction and magnitude uncertainty is absorbed, thereby reflecting pure discretization and boundary-condition fidelity.
This six-degree-of-freedom basis matches the force-plus-moment load that SimJEB applies through an RBE3 reference point; a force-only (three-DOF) basis cannot reproduce the torsional case.
Fitting a displacement field of tens of thousands of nodal degrees of freedom with only six coefficients ($6 \ll 3N$) is far from an over-parameterized fit.

\begin{table*}[tp]
\centering
\caption{Accuracy of the automated CAE pipeline on the reference SimJEB brackets, where ground truth exists. \emph{Top}: node-level displacement agreement from a six-DOF (three-force, three-moment) least-squares fit, over 289 of the 294 validated brackets (five near-zero-deformation cases excluded). \emph{Bottom}: agreement of the computed mass label with the CAD ground truth.}
\label{tab:fea_validation}
\begin{tabular}{lcccc}
\toprule
& \textbf{Vertical} & \textbf{Horizontal} & \textbf{Diagonal} & \textbf{Torsional} \\
\midrule
Median $R^2$ & 0.94 & 0.94 & 0.88 & 0.91 \\
Interquartile range & 0.88--0.97 & 0.89--0.97 & 0.76--0.96 & 0.86--0.94 \\
Cases with $R^2\!\geq\!0.8$ (\%) & 88 & 90 & 69 & 91 \\
Best case (\#556) & 0.998 & 0.998 & 0.994 & 0.994 \\
\midrule
\multicolumn{5}{l}{Mass label (mesh volume $\times$ density) vs CAD GT: $R^2{=}0.9999$, median $|$err$|{=}0.06\%$, $99\%$ within $1\%$}\\
\bottomrule
\end{tabular}
\end{table*}

\begin{figure*}[tp]
    \centering
    \includegraphics[width=\textwidth]{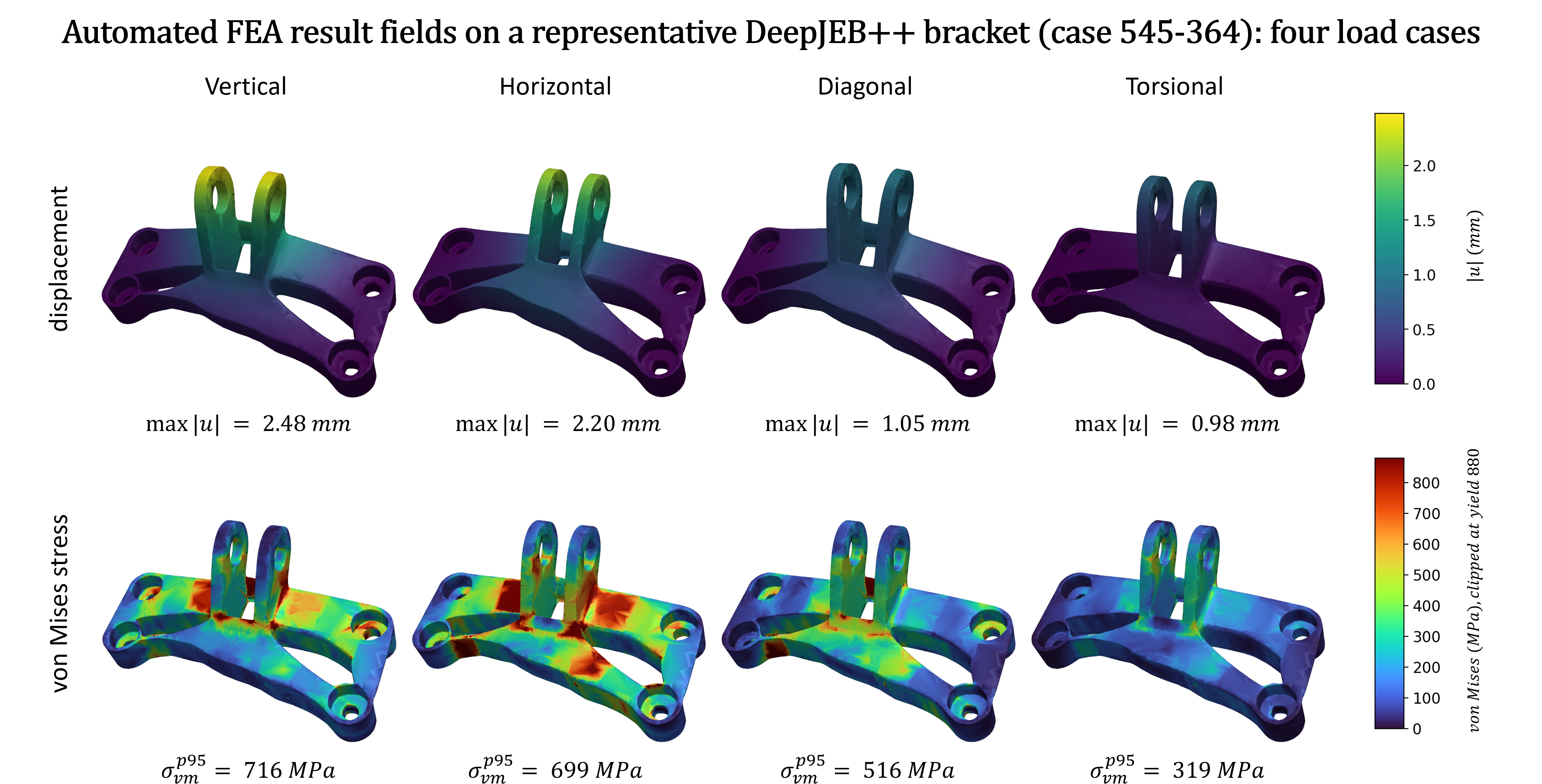}
    \caption{Automated FEA result fields for a representative deployable \ours{} bracket (case~545-364) under the four load cases (columns). \emph{Top}: displacement magnitude $|u|$ on the deformed shape ($9\times$ exaggerated). \emph{Bottom}: nodal von Mises stress, clipped at the $903$\,MPa Ti--6Al--4V yield.}
    \label{fig:fea_fields}
\end{figure*}

\begin{figure*}[tp]
    \centering
    \includegraphics[width=\textwidth]{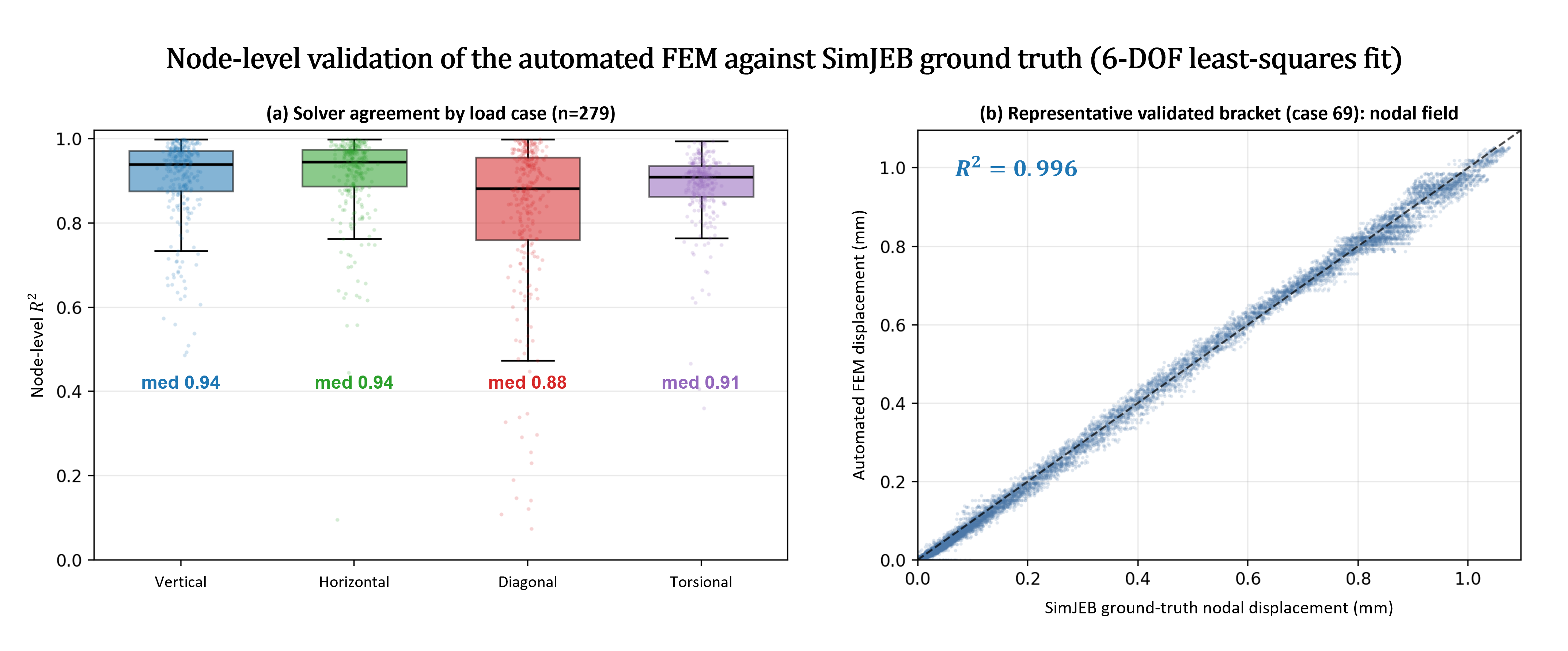}
    \caption{Node-level validation of the automated FEM against the SimJEB ground truth (Section~\ref{sec:exp_stage3}). (a)~Per-bracket $R^2$ distribution by load case over the 289 validated brackets, medians annotated. (b)~Per-node displacement for a representative bracket (case~69) against the reference field.}
    \label{fig:fea_validation}
\end{figure*}

Table~\ref{tab:fea_validation} and Figure~\ref{fig:fea_validation} report the node-level $R^2$ across the four load cases: the medians fall between $0.88$ and $0.94$, the bulk of brackets exceed $0.8$ (88--91\% for the in-plane and torsional loads, 69\% for the harder diagonal case), and the best-conditioned bracket (\#556) reaches $0.99$--$1.00$ on every load.
Figure~\ref{fig:fea_validation}(b) shows this agreement at the node level for a representative bracket (case~69), whose per-node displacements track the reference along the identity line ($R^2=0.996$, slope~$1.00$, $n=13{,}204$ nodes).
A small number of low-deformation cases (peak displacement $<0.3$\,mm) fall near the solver's numerical noise floor and yield unstable $R^2$, so we report the median rather than the mean.
These results indicate that, up to load identification, the automated pipeline's finite element solution is in close agreement with the reference SimJEB simulation.
The residual gap from unity is attributable to the bolt-constraint boundary condition---our full clamp versus the reference's RBE2 bolt spider, which permits rotation about the bolt axis (distinct from the RBE3 point used for load application above)---and to mesh decimation.
The mass label, computed as mesh volume times material density, is essentially exact: it matches the SimJEB CAD ground-truth mass with $R^2=0.9999$ and a median relative error of $0.06\%$, with $99\%$ of the 279 single-body meshes within $1\%$ (Table~\ref{tab:fea_validation}).
Beyond these aggregate validation metrics, Figure~\ref{fig:fea_fields} shows the spatial displacement and von Mises stress fields that the automated pipeline produces for a representative deployable bracket (case~545-364, whose four-load responses lie on the deployable-set medians) under the four load cases.
The pipeline recovers physically consistent load paths: the loaded clevis and the lower web carry the in-plane vertical, horizontal, and diagonal cases, whereas the torsional case engages the part more uniformly at lower amplitude (peak $|u|=2.48/2.20/1.05/0.98$\,mm and $\sigma_{\text{vm}}^{p95}=716/699/516/319$\,MPa for the vertical/horizontal/diagonal/torsional cases, respectively).
These per-load fields constitute the dataset labels.
The fields are re-solved on a finer mesh for visualization, so their absolute magnitudes are mesh-resolution dependent; node-level fidelity is therefore established separately against the ground truth (Fig.~\ref{fig:fea_validation}), and dataset-level comparisons use self-normalized responses (Fig.~\ref{fig:response_dist}).

\subsection{Dataset Statistics}
\label{sec:dataset_stats}

Table~\ref{tab:dataset_comparison} compares \ours{} with existing jet engine bracket datasets.

\begin{table*}[tp]
\centering
\caption{Comparison of jet engine bracket datasets. Seed counts are as reported by each source: DeepJEB~\citep{hong2025deepjeb} used 263 SimJEB brackets, whereas \ours{} uses 380 of the 381.}
\label{tab:dataset_comparison}
\begin{tabular}{lccc}
\toprule
& \textbf{SimJEB} & \textbf{DeepJEB} & \textbf{\ours{}} \\
\midrule
Seed data & --- & 263 (from SimJEB) & 380 (from SimJEB) \\
Total 3D samples & 381 & 2{,}138 & \textbf{15{,}360} \\
Simulation labels & FEA (manual setup) & FEA (automated) & \textbf{FEA (automated BC + CAE)} \\
Expansion ratio & 1$\times$ & 5.6$\times$ & \textbf{40$\times$} \\
Generation method & Hand-designed & DeepSDF interpolation & \textbf{Foundation model transfer} \\
2D pretraining & None & None & \textbf{Stable Diffusion (1B+)} \\
3D pretraining & None & None & \textbf{TRELLIS (500K+ shapes)} \\
Quality filtering & Manual + IQR & Min Jacobian + IQR & \textbf{VLM-based classifier} \\
\bottomrule
\end{tabular}
\end{table*}

\begin{figure}[tbp]
\centering
\includegraphics[width=\columnwidth]{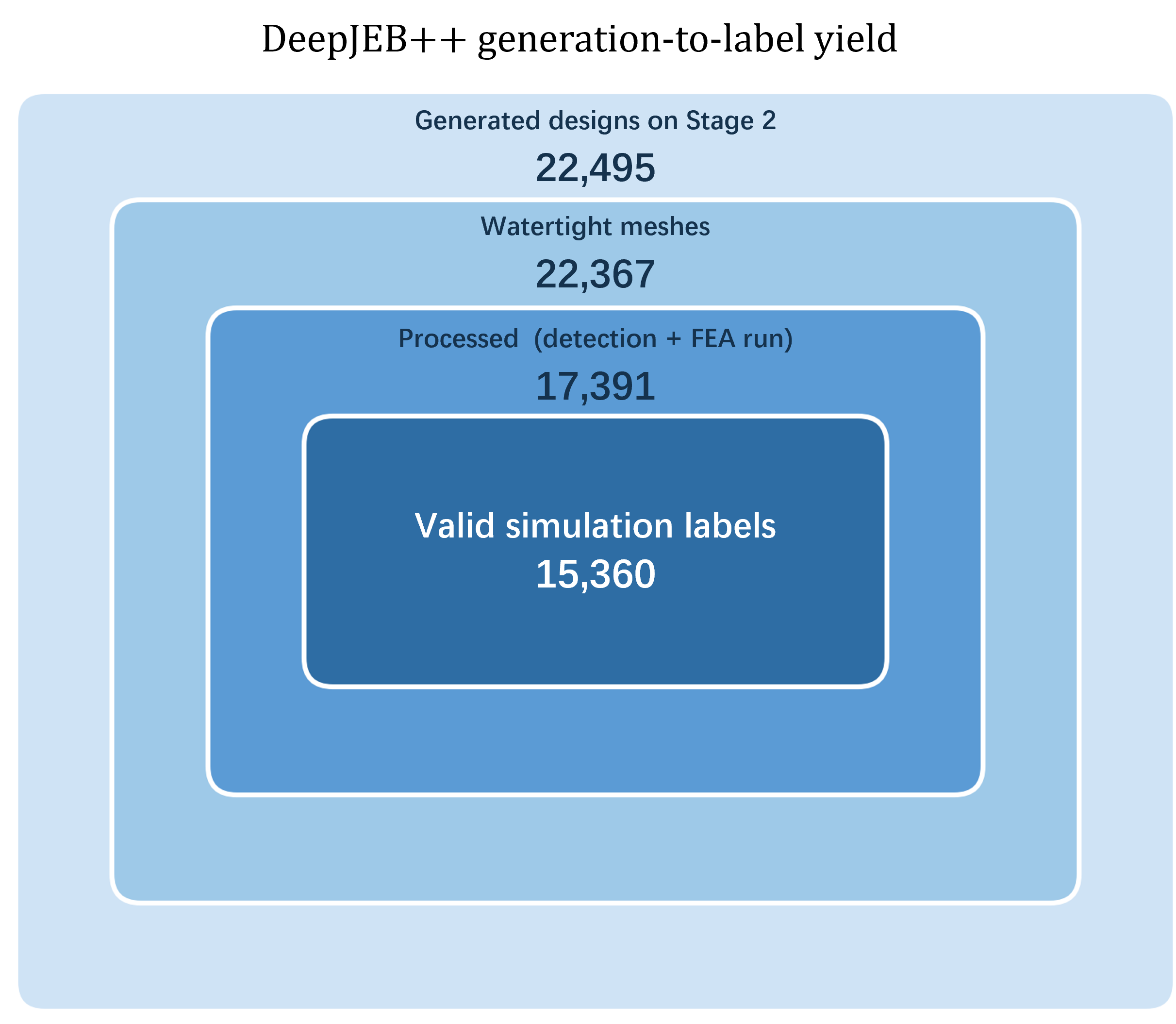}
\caption{Generation-to-label yield of the \ours{} pipeline, from 22{,}495 generated designs to 15{,}360 deployable simulation-labeled brackets. Per-stage counts and losses are given in the text.}
\label{fig:yield}
\end{figure}

We summarize the end-to-end yield of the pipeline (Fig.~\ref{fig:yield}). Of the 22{,}495 generated designs, 22{,}367 produced meshes that entered the automated FEA pipeline, of which 17{,}391 solved successfully. A single-body mesh-quality screen (largest connected-component volume fraction $\geq 0.95$) then removes fragmented and residual multi-body meshes, leaving \textbf{15{,}360 deployable simulation-labeled brackets}---9{,}063 clean single-body meshes plus 6{,}297 from which a minor disconnected speck is removed, after conservatively excluding 39 further cases whose interfaces failed re-detection during automated mass computation. This is a $40\times$ expansion over the 381-bracket SimJEB seed set, produced under a single-GPU budget.
The losses break down as 128 designs lost at mesh generation ($22{,}495\!\to\!22{,}367$), and then, among the $7{,}007$ meshes removed between the watertight-mesh and valid-label stages: no four-bolt interface ($-2{,}428$), the lug--clevis geometric gate ($-1{,}350$), no clevis detected ($-334$), FEM timeout ($-31$), and fragmented mesh/registration / other ($-2{,}864$).

\subsection{Augmentation Coverage in the Latent Space}
\label{sec:coverage}

\begin{figure}[tbp]
\centering
\includegraphics[width=\columnwidth]{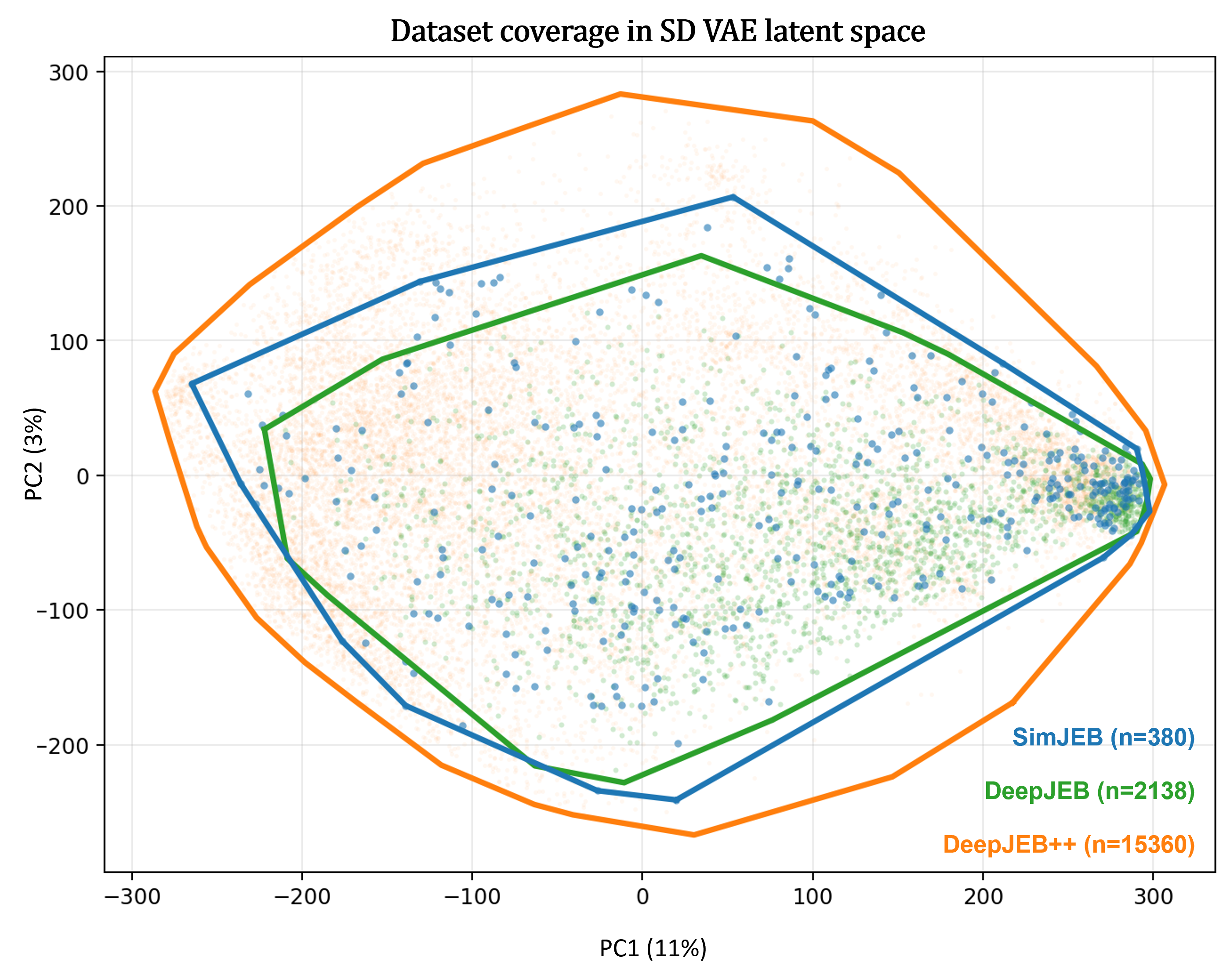}
\caption{Dataset coverage in a pose-normalized SD-VAE appearance space (SimJEB $n{=}380$, DeepJEB $n{=}2{,}138$, \ours{} $n{=}15{,}360$; PCA projection of the appearance embeddings). The SimJEB and DeepJEB samples fall almost entirely within the \ours{} (orange) support, indicating that the augmentation covers and modestly extends the manifold of both while staying on-distribution; the larger \ours{} hull partly reflects its much larger sample size (see text).}
\label{fig:coverage}
\end{figure}

Finally, we examine how the augmented data relate to the seed set and to the prior DeepJEB dataset.
We voxelize every bracket to a common $64^3$ grid, align each shape to a canonical pose by its principal axes (so that orientation does not confound the comparison), and render it from the diagonal three-quarter viewpoint used for single-view generation (the best-performing view in Fig.~\ref{fig:multiview}); the renders are encoded with the Stable Diffusion VAE encoder and the resulting appearance embeddings are projected to two dimensions by PCA (Fig.~\ref{fig:coverage}).
A useful augmentation should remain on the seed data manifold while broadening it, rather than drifting to a disjoint or degenerate region. We test this in a pose-normalized appearance space using all deployable samples (380 SimJEB, 2{,}138 DeepJEB, and 15{,}360 \ours{}; Fig.~\ref{fig:coverage}).\footnote{This geometry-coverage analysis uses the full 380-bracket seed set; the labeled comparisons in Section~\ref{sec:sj_compare} use the 279 single-body SimJEB brackets that additionally carry valid mass labels.} The three datasets overlap substantially: the SimJEB and DeepJEB samples fall almost entirely within the \ours{} support, so \ours{} covers and modestly extends the manifold spanned by both the seed set and the earlier DeepJEB augmentation rather than collapsing or drifting away.

This coverage reflects valid, on-distribution augmentation rather than unbounded geometric novelty. Part of the larger \ours{} hull follows from its order-of-magnitude larger sample count: under a sample-balanced comparison (equal $n$ per dataset) the three convex-hull areas become comparable, and a nearest-neighbor test keeps the datasets separable above the $0.5$ chance level even as they heavily overlap---consistent with augmented samples that are novel instances yet drawn from the seed distribution. The contribution of \ours{} therefore, rests on augmenting the labeled set at scale with physically consistent labels (Section~\ref{sec:exp_stage3}), not on geometric novelty per se.

\subsection{Comparison with SimJEB: Mass and Structural Response}
\label{sec:sj_compare}

\begin{figure*}[tp]
\centering
\includegraphics[width=\textwidth]{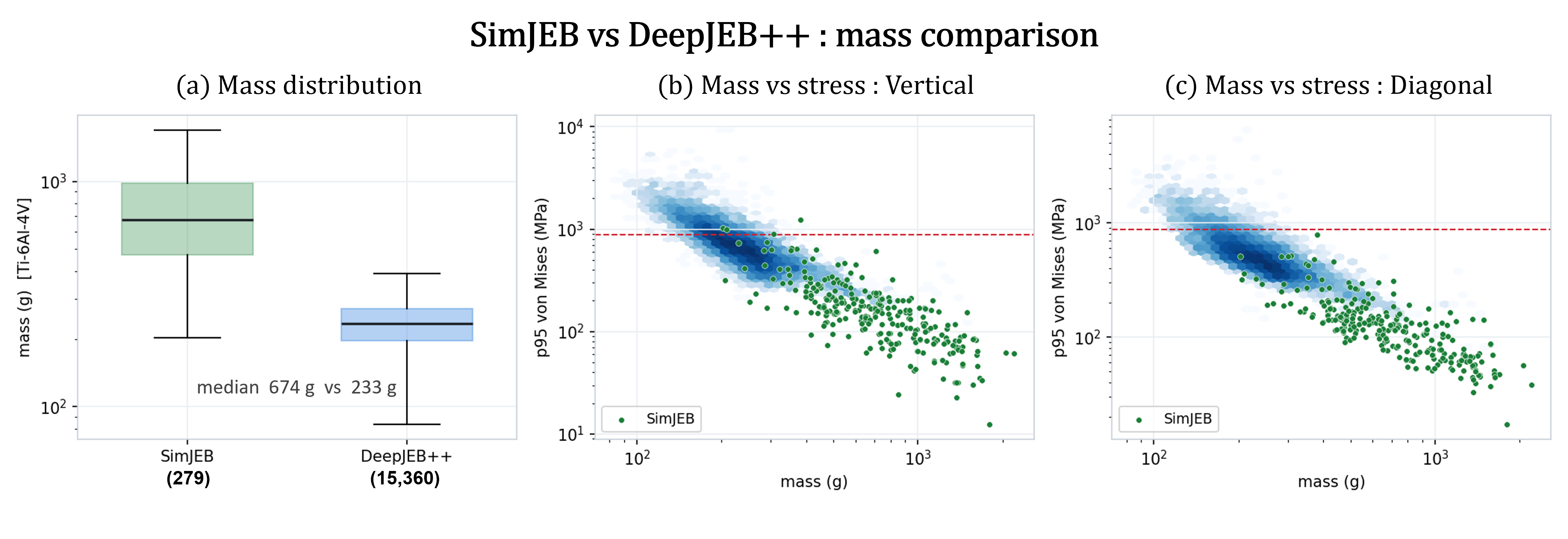}
\caption{Mass comparison between SimJEB ($n{=}279$) and \ours{} ($n{=}15{,}360$ deployable). \emph{(a)~Mass distribution}: part-mass distributions (Ti--6Al--4V), computed from mesh volume and hence mesh-robust. \emph{(b,~c)~Mass vs.\ stress}: $95$th-percentile von Mises stress versus mass for the \emph{(b)}~vertical and \emph{(c)}~diagonal load cases (\ours{} as blue point density, SimJEB as green points; the dashed line marks the $903$\,MPa Ti--6Al--4V yield).}
\label{fig:mass_comp}
\end{figure*}

As a further augmentation-quality check, we test whether the augmented dataset is distributionally consistent with the SimJEB reference---beyond the per-bracket label fidelity established in Section~\ref{sec:exp_stage3}---along two \emph{mesh-robust} axes: part mass and the shape of the structural-response distribution.
Mass is computed directly from the mesh volume and is insensitive to the resolution difference between the datasets---the computed mass matches the CAD ground truth to $R^2 = 0.9999$ (Fig.~\ref{fig:fea_validation}).
The \ours{} brackets are substantially lighter than SimJEB, with a median mass of $233$\,g versus $674$\,g (Fig.~\ref{fig:mass_comp}a), indicating that the generative pipeline explores a thinner, more compliant region of the design space.
The mass--stress scatter (Fig.~\ref{fig:mass_comp}b,c) shows the expected inverse relation---lighter brackets carry higher peak stress---with \ours{} following the same trend as SimJEB while extending the envelope toward lighter, more highly stressed designs.

\begin{figure*}[tp]
\centering
\includegraphics[width=\textwidth]{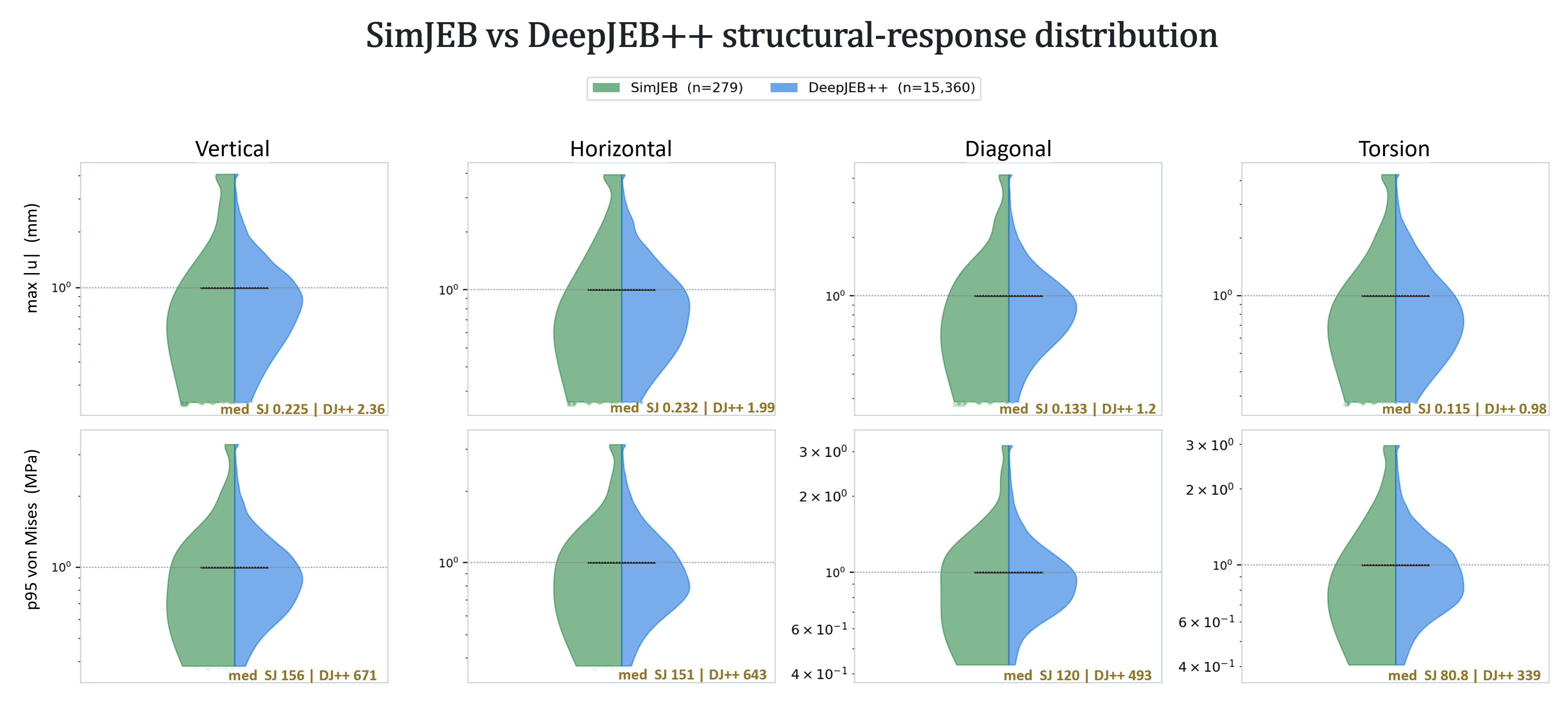}
\caption{Self-normalized structural-response distributions for SimJEB ($n{=}279$) and \ours{} ($n{=}15{,}360$ deployable), per load case (columns) and response (rows: maximum displacement $|u|$, $95$th-percentile von Mises stress). Each sample is divided by its own dataset median (dashed line at $1.0$), making the comparison robust to the mesh-resolution difference; absolute medians (amber) are mesh-confounded and excluded. Node-level fidelity is established separately in Fig.~\ref{fig:fea_validation}.}
\label{fig:response_dist}
\end{figure*}

Absolute displacement and stress magnitudes, by contrast, are mesh-confounded (SimJEB is distributed at ${\sim}50\%$ decimation, whereas \ours{} retains near-full meshes), so we compare only the \emph{shape} of each response distribution after normalizing every sample by its own dataset median (Fig.~\ref{fig:response_dist}).
Across all four load cases, the self-normalized displacement and $95$th-percentile von Mises distributions of \ours{} track those of SimJEB, indicating that the automatically generated labels reproduce the reference response-distribution shape and are physically consistent.

\section{Discussion}
\label{sec:discussion}

\subsection{Foundation Model Knowledge Transfer}
Our results suggest that pretrained foundation models encode geometric priors applicable to this engineering domain.
The fine-tuned Stable Diffusion model's ability to generate bracket-specific designs from a generic text prompt suggests that the model has captured appearance-level regularities of mechanical topology (load paths, structural connectivity) during pretraining on diverse images.
Similarly, TRELLIS's rapid adaptation to the bracket domain (300 training samples) indicates that its pretrained 3D priors (learned from 500K+ shapes) transfer effectively to specialized engineering geometries.

\subsection{Cross-Dimensional Augmentation}
The three-stage 3D $\rightarrow$ 2D $\rightarrow$ 3D $\rightarrow$ simulation pipeline enables us to exploit the asymmetry between 2D and 3D foundation model maturity.
While 2D generative models benefit from billion-scale pretraining datasets, 3D models are limited to hundreds of thousands of shapes.
By augmenting in 2D latent space and reconstructing in 3D, we draw on web-scale 2D priors for generation while securing geometric fidelity through lightweight 3D domain adaptation---a route fundamentally distinct from training a domain-specific implicit generator (e.g., DeepSDF) from scratch.
Because Stages~1--2 adapt general-purpose foundation models rather than a bracket-specific network, the augmentation backbone is domain-agnostic; re-targeting it to a new domain mainly requires adapting the Stage-3 labeling, and demonstrating cross-domain transfer is left to future work.
The automated CAE pipeline in Stage~3 completes the augmented dataset with physics-based performance labels; quantifying the downstream benefit of this augmentation is likewise an important direction for future work.

\subsection{VLM as Engineering Inspector}
The VLM-based quality classifier offers a complementary, semantic approach to automated engineering quality assessment.
Unlike traditional geometric metrics (min Jacobian, aspect ratio), VLM assessment operates on semantic-level understanding of manufacturing feasibility.
The NWN problem we identify highlights an important consideration for deploying language models in technical quality control: the model's tendency to use negation when describing the absence of defects creates systematic scoring errors in similarity-based evaluation.

\subsection{Automated Boundary Condition Recognition}
A key challenge in scaling engineering datasets is that generated geometries must be paired with physically meaningful simulation setups.
The automated boundary condition recognition in Stage~3 addresses this by detecting interface regions (load and bolt points) on generated meshes without manual intervention.
This automation is essential for scaling from hundreds to tens of thousands of simulated designs, as manual boundary condition assignment would be prohibitively expensive at this scale.

\subsection{Limitations and Future Work}
Several limitations should be noted:
\begin{itemize}
    \item VLM classifier accuracy (76.10\%, threshold-tuned; 72.17\% at the fixed default threshold) leaves room for improvement. Fine-tuning VLMs on domain-specific engineering assessment data could substantially improve filtering quality.
    \item The 2D-to-3D conversion introduces some geometric artifacts, particularly in thin-wall regions. Post-processing and mesh repair techniques could mitigate these issues.
    \item The boundary condition recognition module relies on geometric heuristics; learning-based approaches could improve robustness on highly novel topologies.
    \item While our pipeline demonstrates feasibility with jet engine brackets, validation on additional engineering domains (automotive, aerospace, medical devices) would strengthen generalizability claims.
    \item We assess augmentation quality intrinsically (manufacturability, label fidelity, and distributional consistency) but do not yet measure its downstream benefit; quantifying how the augmented data improves surrogate or generative models is an important next step.
    \item Absolute displacement and stress magnitudes are mesh-confounded across datasets---only mass is validated in absolute terms---so the dataset comparison relies on distribution shape rather than absolute values.
\end{itemize}

Future work will focus on: (1) improving boundary condition recognition with learning-based methods for broader domain applicability, (2) AI-driven autonomous design agents that leverage geometry foundation models (GFMs) for rapid design space exploration, and (3) extension to diverse engineering domains beyond brackets.

\section{Conclusion}
\label{sec:conclusion}

We have presented \ours{}, a large-scale 3D engineering dataset of jet engine brackets constructed by leveraging pretrained foundation models under severely constrained resources.
Our three-stage pipeline---2D latent space augmentation with quality-controlled Stable Diffusion interpolation (Stage~1), 3D mesh generation via fine-tuned TRELLIS (Stage~2), and automated boundary condition recognition with CAE simulation (Stage~3)---expands 380 seed designs into 15{,}360 3D geometries with simulation labels, representing a 40$\times$ scale increase over the original SimJEB dataset.

Key technical contributions include: (1) a foundation-model-driven cross-dimensional (2D$\rightarrow$3D) data-augmentation framework for simulation-labeled 3D engineering data, (2) a VLM-based engineering quality filtering framework with the identification and resolution of the NWN problem, (3) domain-specific fine-tuning of a 3D generative foundation model with significant reconstruction quality improvements, and (4) an automated boundary condition recognition and CAE simulation pipeline that produces physics-based performance labels at scale.

By demonstrating that domain-specific GFM pipelines can be constructed with a single GPU in days rather than hundreds of GPUs over weeks, we provide a practical pathway for research labs and small enterprises to build foundation model-driven engineering datasets.
Crucially, because the Stage-1--2 augmentation backbone adapts general-purpose foundation models rather than a generator trained from scratch on the target domain, it is domain-agnostic and---unlike domain-specific implicit approaches rebuilt per domain---can in principle be re-targeted to new domains by adapting the labeling stage, a direction we leave to future work.
The dataset will be publicly released.

\FloatBarrier
\section*{Acknowledgments}
\ifblind
Acknowledgments and funding information have been omitted to preserve anonymity for double-blind review.
\else
This work was supported by grants from the Ministry of Science and ICT (GTL24031-000, N10250154, No.2022-0-00986) and the Ministry of Trade, Industry \& Energy (RS-2025-02317327, RS-2025-25444634).
The authors thank the KAIST Smart Design Lab members for helpful discussions.

\section*{Funding Data}
\begin{itemize}
\item Ministry of Science and ICT (Grant Nos.\ GTL24031-000, N10250154, and 2022-0-00986).
\item Ministry of Trade, Industry \& Energy (Grant Nos.\ RS-2025-02317327 and RS-2025-25444634).
\end{itemize}
\fi

\section*{Conflict of Interest}
The authors declare that they have no conflict of interest.

\section{Licensing, Attribution, and Accessibility}
\label{sec:licensing}

The \ours{} dataset is a publicly available resource designed to facilitate data-driven research in 3D engineering design, surrogate modeling, and structural performance prediction. It comprises 15{,}360 simulation-labeled jet engine brackets generated through the integrated foundation-model pipeline of Section~\ref{sec:method}, pairing each geometry with automatically computed finite-element labels. An overview of the dataset components is provided in Table~\ref{tab:dataset_components}.

\begin{table}[t]
\centering
\footnotesize
\caption{Overview of \ours{} dataset components. The mesh, boundary-condition, and field components are distributed as per-component compressed archives (\texttt{.tar.gz}) that extract to the directories listed below.}
\label{tab:dataset_components}
\begin{tabular}{@{}p{1.35cm}p{2.25cm}p{3.65cm}@{}}
\toprule
\textbf{Category} & \textbf{Path} & \textbf{Description} \\
\midrule
Surface meshes & \texttt{1\_surface\_meshes.tar.gz} & 15{,}360 pose-normalized, watertight triangle meshes (\texttt{.obj}). \\
Boundary conditions & \texttt{2\_boundary\_conditions.tar.gz} & Four-bolt and loaded-clevis interfaces (vertex indices and millimeter frame) per bracket. \\
Per-load fields & \texttt{3\_fea\_fields.tar.gz} & Nodal displacement $|u|$ and von Mises stress fields, four load cases ($2\times4$ per bracket). \\
Scalar labels & \texttt{deepjebpp\_labels.csv} & Per-load maximum displacement and $95$th-percentile von Mises stress. \\
Mass & \texttt{deepjebpp\_labels.csv} & Part mass (enclosed volume $\times$ Ti--6Al--4V density), from the volume-to-mass pipeline validated to $R^2{=}0.9999$ against CAD on the SimJEB reference set. \\
Metadata & \texttt{metadata.json} & Material (Ti--6Al--4V), the four load-case magnitudes/directions, and units. \\
\bottomrule
\end{tabular}
\end{table}

\smallskip\noindent\textbf{Generation Process}\quad
All bracket geometries were synthesized by adapting pretrained 2D and 3D foundation models---Stable Diffusion~v1.5~\citep{rombach2022stablediffusion} and TRELLIS~\citep{xiang2025trellis}---through the three-stage pipeline of Section~\ref{sec:method}. Simulation labels were produced by the automated boundary-condition recognition and linear-elastic finite-element pipeline of Section~\ref{sec:stage3}, and validated against the SimJEB ground truth (Section~\ref{sec:exp_stage3}).

\smallskip\noindent\textbf{Hosting and Compatibility}\quad
The dataset is publicly hosted on the Hugging Face Hub \ifblind(repository URL withheld for double-blind review; available to reviewers upon request through the editorial office)\else at \url{https://huggingface.co/datasets/KAIST-SmartDesignLab/DeepJEB-PP}\fi, released as per-component compressed archives, and is compatible with standard commercial and open-source CAD/CAE platforms as well as Python-based simulation environments. The finite-element volume (tetrahedral) mesh is not distributed explicitly; it is reproducible from the surface mesh using the tetrahedralization settings of Section~\ref{sec:fea_solver}.

\smallskip\noindent\textbf{License}\quad
The \ours{} dataset is a derivative work built upon the SimJEB~\citep{whalen2021simjeb} and DeepJEB~\citep{hong2025deepjeb} datasets, both of which are released under the Open Data Commons Attribution License (ODC-By~v1.0). Consistent with these upstream resources, \ours{} is likewise distributed under the Open Data Commons Attribution License (ODC-By~v1.0), which permits use, modification, and redistribution---including for commercial purposes---provided that proper attribution is given. The underlying foundation models (TRELLIS and Stable Diffusion~v1.5) remain governed by their own permissive licenses.

\smallskip\noindent\textbf{Availability and Maintenance}\quad
The dataset will remain publicly accessible for at least ten years following its release. Any updates, revisions, or corrections will be announced with appropriate version control and a revised access link. A README file is included to assist users in navigating the dataset.

\bibliographystyle{plainnat}

\begin{thebibliography}{99}

\bibitem{oh2019deep}
Oh, S., Jung, Y., Kim, S., Lee, I., and Kang, N.
\newblock Deep generative design: Integration of topology optimization and generative models.
\newblock \emph{ASME J. Mech. Des.}, 141(11):111405, 2019.


\bibitem{regenwetter2022deep}
Regenwetter, L., Nobari, A.~H., and Ahmed, F.
\newblock Deep generative models in engineering design: A review.
\newblock \emph{ASME J. Mech. Des.}, 144(7):071704, 2022.


\bibitem{nie2021topologygan}
Nie, Z., Lin, T., Jiang, H., and Kara, L.~B.
\newblock TopologyGAN: Topology optimization using generative adversarial networks based on physical fields over the initial domain.
\newblock \emph{ASME J. Mech. Des.}, 143(3):031715, 2021.


\bibitem{cunningham2019investigation}
Cunningham, J.~D., Simpson, T.~W., and Tucker, C.~S.
\newblock An investigation of surrogate models for efficient performance-based decoding of 3D point clouds.
\newblock \emph{ASME J. Mech. Des.}, 141(12):121401, 2019.


\bibitem{pfaff2021meshgraphnets}
Pfaff, T., Fortunato, M., Sanchez-Gonzalez, A., and Battaglia, P.~W.
\newblock Learning mesh-based simulation with graph networks.
\newblock \emph{ICLR}, 2021.


\bibitem{whalen2021simjeb}
Whalen, E., Beyene, A., and Mueller, C.
\newblock SimJEB: Simulated jet engine bracket dataset.
\newblock \emph{Computer Graphics Forum}, 40(5):9--17, 2021.


\bibitem{hong2025deepjeb}
Hong, S., Kwon, Y., Shin, D., Park, J., and Kang, N.
\newblock DeepJEB: 3D deep learning-based synthetic jet engine bracket dataset.
\newblock \emph{ASME J. Mech. Des.}, 147(4):041703, 2025.


\bibitem{park2019deepsdf}
Park, J.~J., Florence, P., Straub, J., Newcombe, R., and Lovegrove, S.
\newblock DeepSDF: Learning continuous signed distance functions for shape representation.
\newblock \emph{CVPR}, 2019.


\bibitem{physicsx2024gfm}
PhysicsX.
\newblock Building beyond human imagination with foundation models for geometry and physics.
\newblock \url{https://www.physicsx.ai/newsroom/}, 2024.


\bibitem{chang2015shapenet}
Chang, A.~X., Funkhouser, T., Guibas, L., et~al.
\newblock ShapeNet: An information-rich 3D model repository.
\newblock \emph{arXiv preprint arXiv:1512.03012}, 2015.


\bibitem{koch2019abc}
Koch, S., Matveev, A., Jiang, Z., et~al.
\newblock ABC: A big CAD model dataset for geometric deep learning.
\newblock \emph{CVPR}, 2019.


\bibitem{regenwetter2023framed}
Regenwetter, L., Weaver, C., and Ahmed, F.
\newblock FRAMED: An AutoML approach for structural performance prediction of bicycle frames.
\newblock \emph{Computer-Aided Design}, 156, 2023.


\bibitem{regenwetter2022biked}
Regenwetter, L., Curry, B., and Ahmed, F.
\newblock BIKED: A dataset for computational bicycle design with machine learning benchmarks.
\newblock \emph{ASME J. Mech. Des.}, 144(3):031706, 2022.


\bibitem{bagazinski2023shipd}
Bagazinski, N.~J. and Ahmed, F.
\newblock Ship-D: Ship hull dataset for design optimization using machine learning.
\newblock \emph{IDETC/CIE}, 2023.


\bibitem{elrefaie2024drivaernetpp}
Elrefaie, M., Morar, F., Dai, A., and Ahmed, F.
\newblock DrivAerNet++: A large-scale multimodal car dataset with CFD simulations and deep learning benchmarks.
\newblock \emph{NeurIPS}, 2024.


\bibitem{cobb2023aircraftverse}
Cobb, A.~D., et~al.
\newblock AircraftVerse: A large-scale multimodal dataset of aerial vehicle designs.
\newblock \emph{NeurIPS Datasets and Benchmarks Track}, 2023.


\bibitem{nichol2022pointe}
Nichol, A., Jun, H., Dhariwal, P., Mishkin, P., and Chen, M.
\newblock Point-E: A system for generating 3D point clouds from complex prompts.
\newblock \emph{arXiv:2212.08751}, 2022.


\bibitem{liu2023zero123}
Liu, R., Wu, R., Van~Hoorick, B., Tokmakov, P., Zakharov, S., and Vondrick, C.
\newblock Zero-1-to-3: Zero-shot one image to 3D object.
\newblock \emph{ICCV}, 2023.


\bibitem{poole2023dreamfusion}
Poole, B., Jain, A., Barron, J.~T., and Mildenhall, B.
\newblock DreamFusion: Text-to-3D using 2D diffusion.
\newblock \emph{ICLR}, 2023.


\bibitem{xiang2025trellis}
Xiang, J., Lv, Z., Xu, S., Deng, Y., Wang, R., Zhang, B., Chen, D., Tong, X., and Yang, J.
\newblock Structured 3D latents for scalable and versatile 3D generation (TRELLIS).
\newblock \emph{CVPR}, 2025.


\bibitem{mildenhall2020nerf}
Mildenhall, B., Srinivasan, P.~P., Tancik, M., Barron, J.~T., Ramamoorthi, R., and Ng, R.
\newblock NeRF: Representing scenes as neural radiance fields for view synthesis.
\newblock \emph{ECCV}, 2020.


\bibitem{kerbl2023gaussian}
Kerbl, B., Kopanas, G., Leimk\"uhler, T., and Drettakis, G.
\newblock 3D Gaussian splatting for real-time radiance field rendering.
\newblock \emph{ACM Trans. Graph.}, 42(4), 2023.


\bibitem{liu2023rectifiedflow}
Liu, X., Gong, C., and Liu, Q.
\newblock Flow straight and fast: Learning to generate and transfer data with rectified flow.
\newblock \emph{ICLR}, 2023.


\bibitem{hong2024lrm}
Hong, Y., Zhang, K., Gu, J., Bi, S., Zhou, Y., Liu, D., Liu, F., Sunkavalli, K., Bui, T., and Tan, H.
\newblock LRM: Large reconstruction model for single image to 3D.
\newblock \emph{ICLR}, 2024.


\bibitem{liu2023one2345}
Liu, M., Xu, C., Jin, H., Chen, L., Varma~T, M., Xu, Z., and Su, H.
\newblock One-2-3-45: Any single image to 3D mesh in 45 seconds without per-shape optimization.
\newblock \emph{NeurIPS}, 2023.


\bibitem{long2024wonder3d}
Long, X., Guo, Y.-C., Lin, C., Liu, Y., Dou, Z., Liu, L., Ma, Y., Zhang, S.-H., Habermann, M., Theobalt, C., and Wang, W.
\newblock Wonder3D: Single image to 3D using cross-domain diffusion.
\newblock \emph{CVPR}, 2024.


\bibitem{xu2024instantmesh}
Xu, J., Cheng, W., Gao, Y., Wang, X., Gao, S., and Shan, Y.
\newblock InstantMesh: Efficient 3D mesh generation from a single image with sparse-view large reconstruction models.
\newblock \emph{arXiv:2404.07191}, 2024.


\bibitem{li2025triposg}
Li, Y., Zou, Z.-X., Liu, Z., Wang, D., Liang, Y., Yu, Z., Liu, X., Guo, Y.-C., Liang, D., Ouyang, W., and Cao, Y.-P.
\newblock TripoSG: High-fidelity 3D shape synthesis using large-scale rectified flow models.
\newblock \emph{arXiv:2502.06608}, 2025.


\bibitem{hunyuan3d}
Tencent Hunyuan3D Team.
\newblock Hunyuan3D 2.0: Scaling diffusion models for high-resolution textured 3D assets generation.
\newblock \emph{arXiv:2501.12202}, 2025.


\bibitem{chen2021padgan}
Chen, W. and Ahmed, F.
\newblock PaDGAN: Learning to generate high-quality novel designs.
\newblock \emph{ASME J. Mech. Des.}, 143(3):031703, 2021.


\bibitem{nobari2021pcdgan}
Heyrani~Nobari, A., Chen, W., and Ahmed, F.
\newblock PcDGAN: A continuous conditional diverse generative adversarial network for inverse design.
\newblock \emph{KDD}, 2021.


\bibitem{shu20203d}
Shu, D., Cunningham, J., Stump, G., et~al.
\newblock 3D design using generative adversarial networks and physics-based validation.
\newblock \emph{ASME J. Mech. Des.}, 142(7):071701, 2020.


\bibitem{wang2022ihgan}
Wang, J., Chen, W.~W., Da, D., Fuge, M., and Rai, R.
\newblock IH-GAN: A conditional generative model for implicit surface-based inverse design of cellular structures.
\newblock \emph{CMAME}, 396:115060, 2022.


\bibitem{maze2023diffusion}
Maz\'e, F. and Ahmed, F.
\newblock Diffusion models beat GANs on topology optimization.
\newblock \emph{Proc. AAAI Conf. Artif. Intell.}, 37(8):9108--9116, 2023.


\bibitem{radford2021clip}
Radford, A., Kim, J.~W., Hallacy, C., et~al.
\newblock Learning transferable visual models from natural language supervision.
\newblock \emph{ICML}, 2021.


\bibitem{li2023blip2}
Li, J., Li, D., Savarese, S., and Hoi, S.
\newblock BLIP-2: Bootstrapping language-image pre-training with frozen image encoders and large language models.
\newblock \emph{arXiv:2301.12597}, 2023.


\bibitem{liu2023llava}
Liu, H., Li, C., Wu, Q., and Lee, Y.~J.
\newblock Visual instruction tuning.
\newblock \emph{arXiv:2304.08485}, 2023.


\bibitem{rombach2022stablediffusion}
Rombach, R., Blattmann, A., Lorenz, D., Esser, P., and Ommer, B.
\newblock High-resolution image synthesis with latent diffusion models.
\newblock \emph{CVPR}, 2022.


\bibitem{ho2020ddpm}
Ho, J., Jain, A., and Abbeel, P.
\newblock Denoising diffusion probabilistic models.
\newblock \emph{NeurIPS}, 2020.


\bibitem{wang2023interpolating}
Wang, C. and Golland, P.
\newblock Interpolating between images with diffusion models.
\newblock \emph{ICML 2023 Workshop on Challenges in Deployable Generative AI}, 2023.


\bibitem{caron2021dino}
Caron, M., Touvron, H., Misra, I., et~al.
\newblock Emerging properties in self-supervised vision transformers.
\newblock \emph{ICCV}, 2021.


\bibitem{kasa1976circle}
Kasa, I.
\newblock A circle fitting procedure and its error analysis.
\newblock \emph{IEEE Trans. Instrum. Meas.}, 25(1):8--14, 1976.


\bibitem{umeyama1991}
Umeyama, S.
\newblock Least-squares estimation of transformation parameters between two point patterns.
\newblock \emph{IEEE Trans. Pattern Anal. Mach. Intell.}, 13(4):376--380, 1991.


\bibitem{zhao2023unipc}
Zhao, W., Bai, L., Rao, Y., Zhou, J., and Lu, J.
\newblock UniPC: A unified predictor-corrector framework for fast sampling of diffusion models.
\newblock \emph{NeurIPS}, 2023.


\bibitem{ho2022cfg}
Ho, J. and Salimans, T.
\newblock Classifier-free diffusion guidance.
\newblock \emph{arXiv:2207.12598}, 2022.


\end{thebibliography}

\end{document}